\documentclass[10pt,conference,a4paper]{IEEEtran}

\usepackage{graphics} 
\usepackage{epsfig} 
\usepackage{graphicx}
\usepackage{subfigure}
\usepackage{amsmath} 
\usepackage{amssymb}  

\def\etal{\emph{et al.}~}

\usepackage{mathtools}

\DeclarePairedDelimiter\floor{\lfloor}{\rfloor}
\newcommand{\norm}[1]{\left\lVert#1\right\rVert}
\usepackage{footnote}
\usepackage{url}
\usepackage{hyperref}
\usepackage{tikz}
\usetikzlibrary{shapes.multipart,matrix,positioning}
\def\drawLineBelowRow#1#2{
  \draw (#2.west|-#2-#1-1.south) -- (#2.east|-#2-#1-1.south);
}

\begin{document}
\urlstyle{stt}
\title{A Deep Multi-Level Network\\ for Saliency Prediction}
\author{\IEEEauthorblockN{Marcella Cornia, Lorenzo Baraldi, Giuseppe Serra and Rita Cucchiara}
\IEEEauthorblockA{Dipartimento di Ingegneria ``Enzo Ferrari''\\
Universit\`a degli Studi di Modena e Reggio Emilia\\
Email: \{name.surname\}@unimore.it}}

\maketitle

\begin{abstract}
This paper presents a novel deep architecture for saliency prediction. Current state of the art models for saliency prediction employ Fully Convolutional networks that  perform a non-linear combination of features extracted from the last convolutional layer to predict saliency maps. We propose an architecture which, instead, combines features extracted at different levels of a Convolutional Neural Network (CNN). Our model is composed of three main blocks: a feature extraction CNN, a feature encoding network, that weights low and high level feature maps, and a prior learning network. 
We compare our solution with state of the art saliency models on two public benchmarks datasets. Results show that our model outperforms under all evaluation metrics on the SALICON dataset, which is currently the largest public dataset for saliency prediction, and achieves competitive results on the MIT300 benchmark. Code is available at \texttt{\url{https://github.com/marcellacornia/mlnet}}. 
\end{abstract}


\section{Introduction}
When human observers look at an image, effective attentional mechanisms attract their gazes on salient regions which have distinctive variations in visual stimuli. 
Emulating such ability has been studied for more than 80 years by neuroscientists~\cite{buswell1935people} and more recently by computer vision researches~\cite{itti1998model}.
 
Traditionally, algorithms for saliency prediction focused on identifying the fixation points that human viewers would focus on at first glance. Others have concentrated on highlighting the most important object regions in an image~\cite{Li_2015_CVPR,Yan_2013_CVPR,Jiang_2013_CVPR}. 
We focus on the first type of saliency models, that try to predict eye fixations over an image. 

Inspired by biological studies, researchers have defined hand-crafted and multi-scale features that capture a large spectrum of stimuli: lower-level features (color, texture, contrast)~\cite{hou2007saliency} or higher-level concepts (faces, people, text, horizon)~\cite{cerf2009faces}. However, given the large variety of aspects that can contribute to define visual saliency, it is difficult to design approaches that combine all these hand-tuned factors in an appropriate way.     
 
In recent years, Deep learning techniques have shown impressive results in several vision tasks, such as image classification~\cite{krizhevsky2012imagenet} and semantic segmentation~\cite{long2015fully}. Motivated by these achievements, first attempts to predict saliency map with deep convolutional networks have been performed~\cite{vig2014large,kummerer2014deep}. These solutions suffered from the small amount of training data compared to the ones available in other contexts requiring the usage of limited number of layers or the usage of pretrained architectures generated for other tasks. The recent publication of the large dataset SALICON~\cite{huang2015salicon}, collected thanks to crowd-sourcing techniques, allows researches to increase the number of convolutional layers reducing the overfitting risk~\cite{kruthiventi2015deepfix,cPan}.

Moreover, it is well known that when observers view complex scenes presented on computer monitors, there is a strong tendency to look more frequently around the center of the scene than around the periphery~\cite{tatler2007central}. This has been exploited in past works on saliency prediction, by incorporating hand-crafted priors into saliency maps, or by learning the relative contribution of different priors.

In this paper we present a deep learning architecture for predicting saliency maps, which exploits multi-level features extracted from a CNN, while still being trainable end-to-end. In contrast to the current trend, we let the network learn its own prior from training data. A new loss function is also designed to train the proposed network and to tackle the imbalance problem of saliency maps.

\section{Related Work}
In the last decade saliency prediction has been widely studied. The seminal works by Koch and Ullman~\cite{koch1987shifts} and Itti~\etal~\cite{itti1998model} introduced a biologically-plausible architecture for saliency detection that extracts multi-scale image features based on color, intensity and orientation. Later, Hou and Zange~\cite{hou2007saliency} proposed a method that analyzes the log spectrum of each image and obtains the spectral residual, which allows the estimation of the behavior of pre-attentive visual search. Torralba~\etal~\cite{torralba2006contextual} show how global contextual information can improve the prediction of observers' eye movements in real-world scenes.

Similarly, Goferman~\etal~\cite{goferman2012context} present a technique which aims at identifying salient regions that are distinctive with respect to both their local and global surroundings. Similarly to Cerf~\etal~\cite{cerf2009faces}, Judd ~\etal~\cite{judd2009learning} propose an approach that combines low-level features (color, orientation and intensity) with high-level semantic information (i.e. the location of faces, cars and text) and show that this solution significantly improves the ability to predict eye fixations. In general, these approaches presented hand-tuned features or trained specific higher-level classifiers.

\begin{figure*}[t!]
\centering
\includegraphics[width=0.85\textwidth]{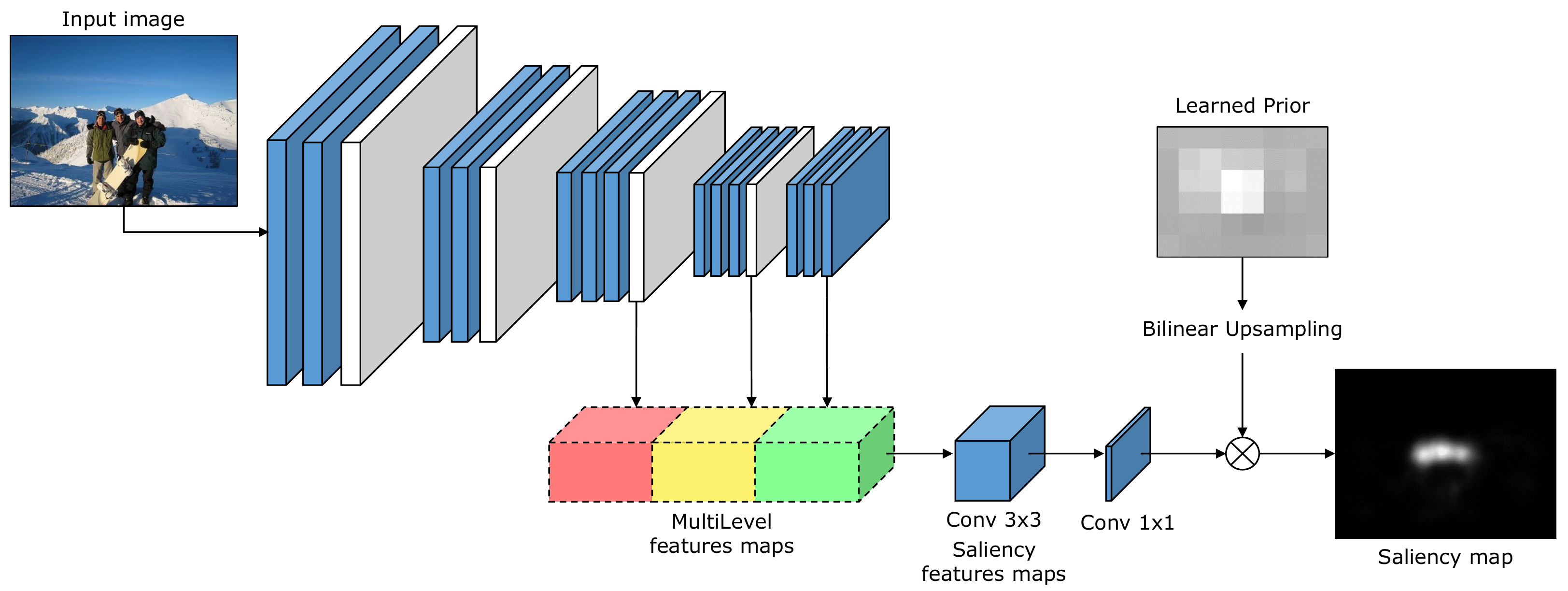} 
\caption{Overview of our model (see Section~\ref{sec:method} for details). A CNN is used to compute low and high level features from the input image. Extracted features maps are then fed to an Encoding network, which learns a feature weighting function to generate saliency-specific feature maps. A prior image is also learned and applied to the predicted saliency map.}
\label{fig:model}
\end{figure*}

A first attempt to model saliency with deep convolutional networks (DCNs) has been recently proposed by Vig~\etal~\cite{vig2014large}. They propose Ensembles of Deep Networks (eDN), a CNN with three layers. Since the amount of data available to learn saliency is generally limited, this architecture cannot scale to outperform the current state-of-the art.

To address this issue, K\"ummerer~\etal~\cite{kummerer2014deep} present a way of reusing existing neural networks, trained for image classification, to predict fixation maps. In particular, they present Deep Gaze, a neural network that uses the well-known AlexNet~\cite{krizhevsky2012imagenet} architecture to generate a high dimensional feature space, which is used to create a saliency map. Similarly, Huang~\etal~\cite{huang2015salicon} propose an architecture that integrates saliency prediction into DCNs pretrained for object recognition (AlexNet~\cite{krizhevsky2012imagenet}, VGG-16~\cite{Simonyan14c} and GoogLeNet~\cite{szegedy2015going}). The key component is a fine-tuning of DNNs weights with an objective function based on the saliency evaluation metrics, such as Normalized Scanpath Saliency (NSS), Similarity and KL-Divergence.

Liu~\etal~\cite{liu2015predicting} propose a multi-resolution CNN that is trained from image regions centered on fixation and non-fixation locations at multi-scales. Recently, Srinivas~\etal\cite{kruthiventi2015deepfix} presented DeepFix, in which they introduce Location Biased Convolution filters that allow the network to exploit location dependent patterns. Pan~\etal~\cite{cPan} present two different architectures: a shallow convnet trained from scratch and a deep convnet that uses parameters previous learned on the ILSVRC-12 dataset.

What the majority of these approaches share is the use of Fully Convolutional Networks which are trained to predict saliency map from a non-linear combination of high level features, extracted from the last convolutional layer. Our architecture learns how to weight features coming from different levels of a CNN, and demonstrates the effectiveness of using medium level features.

\section{Proposed Approach}
\label{sec:method}
Our saliency model is composed by three main parts: given an input image, a CNN extracts low, medium and high level features; then, an encoding network builds saliency-specific features and produces a temporary saliency map. A prior is then learned and applied to produce the final saliency prediction. Figure~\ref{fig:model} reports a summary of the architecture.

\textbf{Feature extraction network}~The first component of our architecture is a Fully Convolutional network with 13 layers, which takes the input image and produces features maps for the encoding network.

We build our architecture on the popular 16 layers model from VGG~\cite{Simonyan14c}, which is well known for its elegance and simplicity, and at the same time yields nearly state of the art results in image classification and good generalization properties. However, like any standard CNN architecture, it has the significant disadvantage of reducing the size of feature maps at higher levels with respect to the input size. This is due to the presence of spatial pooling layers which have a stride greater than one: the output of each layer is a three-dimensional tensor with shape $k \times \floor*{\frac{H}{f}} \times \floor*{\frac{W}{f}}$, where $k$ is the number of filters of the layer, and $f$ is the downsampling factor of that stage of the network. In the VGG-16 model there are five max-pooling stages with kernel size $k=2$ and stride $2$. Given an input image with size $W\times H$, the output feature map has size $ \floor*{\frac{W}{2^5}} \times \floor*{\frac{H}{2^5}}$, thus a Fully Convolutional model built upon the VGG-16 would output a saliency map rescaled by a factor of 32.

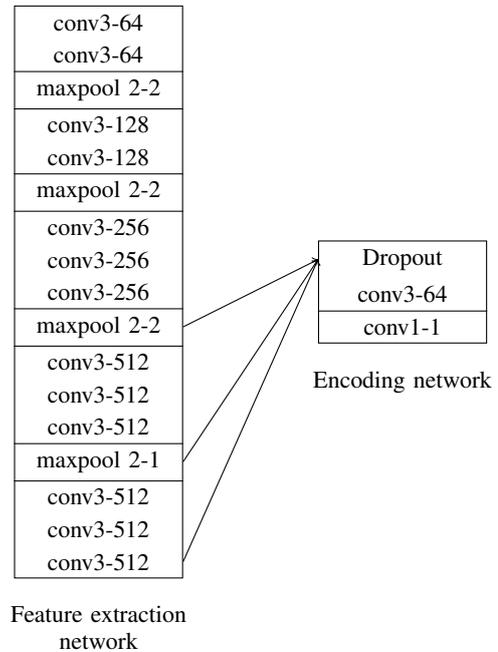
\begin{figure}[tb]
  \centering
  \small
  \begin{tikzpicture}
  \matrix (A)  [matrix of nodes,nodes={minimum size=.40cm, text width=2cm,align=center,inner sep=3pt},draw,inner sep=0pt] at (0,1)
  {
  conv3-64\\
  conv3-64\\
  maxpool 2-2\\
  conv3-128\\
  conv3-128\\
  maxpool 2-2\\  
  conv3-256\\
  conv3-256\\ 
  conv3-256\\  
  maxpool 2-2\\  
  conv3-512\\
  conv3-512\\
  conv3-512\\  
  maxpool 2-1\\    
  conv3-512\\
  conv3-512\\
  conv3-512\\  
  };
  \drawLineBelowRow{2}{A}
  \drawLineBelowRow{3}{A}
  \drawLineBelowRow{5}{A}
  \drawLineBelowRow{6}{A}
  \drawLineBelowRow{9}{A}
  \drawLineBelowRow{10}{A}
  \drawLineBelowRow{13}{A}
  \drawLineBelowRow{14}{A}
  \matrix (B) [matrix of nodes,nodes={minimum size=.40cm, text width=2cm,align=center,inner sep=3pt},draw,inner sep=0pt] at (4,1)
  {
  Dropout\\
  conv3-64\\
  conv1-1\\
  };
  \drawLineBelowRow{2}{B}

  \draw[->] (A-10-1.east) -- (B-1-1.west);
  \draw[->] (A-14-1.east) -- (B-1-1.west);
  \draw[->] (A-17-1.east) -- (B-1-1.west);

  \node (C) [below= .25cm of A,text width=3cm,align=center]{Feature extraction network};
  \node (D) [below= .25cm of B,text width=3cm,align=center]{Encoding network};
  
%
  \end{tikzpicture}
  
  \caption{Architecture of the proposed networks. The convolutional layer parameters are denoted as ``conv\textless receptive field size\textgreater -\textless number of channels\textgreater''. The ReLU activation function is not shown for brevity.}  
 \label{fig:architecture}
\end{figure}

To limit this rescaling phenomenon, we remove the last pooling stage and decrease the stride of the last but one, while keeping unchanged its stride. This way, the output feature map of our feature extraction network are rescaled by a factor of 8 with respect to the input image. In the following, we will refer to the output size of the feature extraction network as $w \times h$, with $w = \floor*{\frac{W}{8}}$ and $h = \floor*{\frac{H}{8}}$. For reference, a complete description of the network is reported in Figure~\ref{fig:architecture}.

\textbf{Encoding network}~We take feature maps at three different locations: the output of the third pooling layer (which contains 256 feature maps), that of the last pooling layer (512 feature maps), and the output of the last convolutional layer (512 feature maps). In the following, we will call these maps, respectively, \texttt{conv3}, \texttt{conv4} and \texttt{conv5}, since they come from the third, fourth and fifth convolutional stage of the network. They all share the same spatial size, and are concatenated to form a tensor with 1280 channels, which is fed to a Dropout layer with retain probability 0.5, to improve generalization. A convolutional layer then learns 64 saliency-specific feature maps with a $3\times 3$ kernel. A final $1\times 1$ convolution learns to weight the importance of each saliency feature map to produce the final predicted feature map.

\textbf{Prior learning}~Instead of using pre-defined priors as done in the past, we let the network learn its own custom prior. To this end, we learn a coarse $w'\times h'$ mask (with $w' \ll w$ and $h' \ll h$), initialized to one, upsample and apply it to the predicted saliency map with pixel-wise multiplication.

Given the learned prior $U$ with shape $w'\times h'$, we interpolate the pixels of $U$ to produce an output prior map $V$ of size $w \times h$. We compute a sampling grid $G$ of shape $w' \times h'$ associating each element of $U$ with real-valued coordinates into $V$. If $G_{i,j} = (x_{i,j} , y_{i,j} )$ then $U_{i,j}$ should be equal to $V$ at $(x_{i,j} , y_{i,j} )$; however since $(x_{i,j} , y_{i,j})$ are real-valued, we convolve with a sampling kernel and set
\begin{equation}
V_{x,y} = \sum_{i=1}^{w'} \sum_{j=1}^{h'} U_{i,j} k_x(x-x_{i,j})k_y(y-y_{i,j})
\end{equation}
where $k_x(\cdot)$ and $k_y(\cdot)$ are bilinear kernels, corresponding to $k_x(d) = \max\left(0,\frac{w}{w'}-|d|\right)$ and $k_y(d) = \max\left(0,\frac{h}{h'}-|d|\right)$. $w'$ and $h'$ were set to $\floor*{w/10}$ and $\floor*{h/10}$ in all our tests.

\textbf{Training}~At training time, we randomly sample a minibatch containing $N$ training saliency maps (in our experiments $N=10$), and encourage the network to minimize a loss function through Stochastic Gradient Descent. Our loss function is inspired by three objectives: predictions should be pixelwise similar to ground truth maps, therefore a square error loss $\| \phi(\mathbf{x}_i) - \mathbf{y} \|^2$ is a reasonable choice. Secondly, predicted maps should be invariant to their maximum, and there is no point in forcing the network to produce values in a given numerical range, so predictions are normalized by their maximum. 
Third, the loss should give the same importance to high and low ground truth values, even though the majority of ground truth pixels are close to zero. For this reason, the deviation between predicted values and ground-truth values $\mathbf{y}_i$ is weighted by a linear function $\alpha-\mathbf{y}_i$, which tends to give more importance to pixels with high ground-truth fixation probability.

The overall loss function is thus
\begin{equation}
L(\mathbf{w}) = \frac{1}{N}\sum_{i=1}^N \norm{ \frac{\frac{\phi(\mathbf{x}_i)}{\max\phi(\mathbf{x}_i)} - \mathbf{y}_i}{\alpha-\mathbf{y}_i}}^2 + \lambda \| \mathbf{1} - U \|^2
\end{equation}
where a $L_2$ regularization term is added to penalize the deviation of the prior mask $U$ from its initial value, thus encouraging the network to adapt to ground truth maps by changing convolutional weights rather than modifying the prior.
Weights for the encoding network are initialized according to~\cite{glorot2010understanding}, and biases are initialized to zero. SGD is applied with Nesterov momentum 0.9, weight decay 0.0005 and learning rate $10^{-3}$. Parameters $\alpha$ and $\lambda$ are respectively set to $1.1$ and $1/(w'\cdot h')$ in all our experiments.

\begin{table*}[tb]
\renewcommand{\arraystretch}{1.05}
\caption{Comparison results on the SALICON test set~\cite{jiang2015salicon}.}
\label{tab:salicon}
\centering
\begin{tabular}{|l|c|c|c|}
\hline
										& CC 			& AUC shuffled 	& AUC Judd		\\ \hline
\textbf{Our method}					& \textbf{0.7430} & \textbf{0.7680}	& \textbf{0.8660} 	\\ \hline
Deep Convnet - Pan \etal~\cite{cPan}	& 0.6220		& 0.7240		& 0.8580		\\ \hline
Shallow Convnet - Pan \etal~\cite{cPan}		& 0.5957		& 0.6698		& 0.8364		\\ \hline
WHU IIP									& 0.4569		& 0.6064		& 0.7923		\\ \hline
Rare 2012 Improved~\cite{riche2013rare2012}	& 0.5108	& 0.6644		& 0.8148		\\ \hline
Xidian									& 0.4811		& 0.6809		& 0.8051		\\ \hline \hline
Baseline: BMS~\cite{zhang2013saliency}	& 0.4268		& 0.6935		& 0.7899		\\ \hline
Baseline: GBVS~\cite{harel2006graph}	& 0.4212		& 0.6303		& 0.7899		\\ \hline
Baseline: Itti~\cite{itti1998model}		& 0.2046		& 0.6101		& 0.6669		\\ \hline
\end{tabular}
\end{table*}

\section{Experimental Evaluation}
\subsection{Experimental setup}
We evaluate the effectiveness of our proposal on two publicly available datasets: SALICON~\cite{jiang2015salicon} and MIT300~\cite{judd2012benchmark}.
SALICON is currently the largest public dataset for saliency prediction, and contains 10,000 training images, 5,000 validation images and 5,000 testing images, taken from the Microsoft CoCo dataset~\cite{lin2014microsoft}. Saliency maps were generated by collecting mouse movements, as a replacement of eye-tracking systems, and authors show an high degree of similarity between their maps and those created from eye-tracking data.

MIT300 is one of the most commonly used datasets for saliency prediction, in spite of its limited size. It consists of 300 natural images, in which Saliency maps have been created from eye-tracking data of 39 observers. Saliency maps are not public available, and predictions must be submitted to the MIT saliency benchmark~\cite{mit-saliency-benchmark} for evaluation. Organizers of the benchmark suggest to use the MIT1003~\cite{judd2009learning} dataset for fine-tuning the model. This includes 1003 images taken from Flickr and LabelMe, generated through eye-tracking data of 15 participants.

\textbf{Evaluation metrics}~Saliency prediction results are usually evaluated with a large variety of metrics: Similarity, Linear Correlation Coefficient (CC), AUC Shuffled, AUC Borji, AUC Judd, Normalized Scanpath Saliency (NSS) and Earth Mover's Distance (EMD). Some of these metrics compare the predicted saliency map with the ground truth map generated from fixation points, while other directly compare the predicted saliency map with fixation points~\cite{riche2013saliency}.

The Similarity metric~\cite{judd2012benchmark} computes the sum of pixel-wise minimums between the predicted saliency map $SM$ and the human eye fixation map $FM$, where $SM$ and $FM$ are supposed to be probability distributions and sum up to one. A similarity score of one indicates that the predicted map is identical to the ground truth one.

The linear correlation coefficient (CC), instead is the Pearson's linear coefficient between $SM$ and $FM$.
It ranges between $-1$ and $1$, and a score close to $-1$ or $1$ indicates a perfect linear relationship between the two maps.

Earth Mover's Distance (EMD) represents the minimal cost to transform the probability distribution of the saliency map $SM$ into the one of the human eye fixations $FM$. Therefore, a larger EMD indicates a larger difference between the two maps.

The Normalized Scanpath Saliency (NSS) metric was defined specifically for the evaluation of saliency models~\cite{peters2005components}. The idea is to quantify the saliency map values at the eye fixation locations and to normalize it whit the saliency map variance
\begin{equation}
NSS(p) = \frac{SM(p) - \mu_{SM}}{\sigma_{SM}}
\end{equation}
where $p$ is the location of one fixation and $SM$ is the saliency map which is normalized to have a zero mean and unit standard deviation. The final NSS score is the average of $NSS(p)$ for all fixations.

Finally, the Area Under the ROC curve (AUC) is one of the most widely used metrics for the evaluation of maps predicted from saliency models. There are several different implementations of this metric. In our experiments we use AUC Judd, AUC Borji and shuffled AUC. The AUC Judd and the AUC Borji choose non-fixation points with a uniform distribution, otherwise shuffled AUC uses human fixations of other images in the dataset as non-fixation distribution. In that way, centered distribution of human fixations of the dataset is taken into account.

\begin{table*}[tb]
\renewcommand{\arraystretch}{1.05}
\caption{Comparison results on the MIT300 test set ~\cite{judd2012benchmark}.}
\label{tab:mit300}
\centering
\begin{tabular}{|l|c|c|c|c|c|c|c|}
\hline
										& Similarity 	& CC 			& AUC shuffled 	& AUC Borji 	& AUC Judd	& NSS	&  EMD 	\\ \hline
Infinite humans  						& 1.00		 	& 1.00 			& 0.80		   	& 0.87			& 0.91		& 3.18	& 0.00	\\ \hline
DeepFix~\cite{kruthiventi2015deepfix}	& \textbf{0.67}	& \textbf{0.78}	& 0.71			& 0.80		& \textbf{0.87}	& \textbf{2.26}	& \textbf{2.04}		\\ \hline
SALICON~\cite{huang2015salicon}			& 0.60			& 0.74			& 0.74			& 0.85		& \textbf{0.87}		& 2.12	& 2.62		\\ \hline
\textbf{Our method}						& 0.59			& 0.67			& 0.70			& 0.75			& 0.85		& 2.05	& 2.63 \\ \hline
Pan \etal - Deep Convnet~\cite{cPan}	& 0.52			& 0.58			& 0.69			& 0.82			& 0.83		& 1.51	& 3.31	\\ \hline
BMS~\cite{zhang2013saliency}			& 0.51			& 0.55			& 0.65			& 0.82			& 0.83		& 1.41	& 3.35	\\ \hline
Deep Gaze 2~\cite{kummerer2014deep}	    & 0.46			& 0.51			& \textbf{0.76} & \textbf{0.86}	& \textbf{0.87} & 1.29	& 4.00		\\ \hline
Mr-CNN~\cite{liu2015predicting}			& 0.48			& 0.48			& 0.69			& 0.75			& 0.79		& 1.37 	& 3.71	\\ \hline
Pan \etal - Shallow Convnet~\cite{cPan}	& 0.46			& 0.53			& 0.64			& 0.78			& 0.80		& 1.47	& 3.99	\\ \hline
GBVS~\cite{harel2006graph}			    & 0.48			& 0.48			& 0.63			& 0.80			& 0.81		& 1.24 	& 3.51	\\ \hline
Rare 2012 Improved~\cite{riche2013rare2012} & 0.46			& 0.42			& 0.67			& 0.75			& 0.77		& 1.34 	& 3.74	\\ \hline
Judd~\cite{judd2009learning}			 & 0.42			& 0.47			& 0.60			& 0.80			& 0.81		& 1.18 	& 4.45	\\ \hline
eDN~\cite{vig2014large}				    & 0.41			& 0.45			& 0.62			& 0.81			& 0.82		& 1.14 	& 4.56	\\ \hline
\end{tabular}
\end{table*}

\subsection{Feature importance analysis}
Our method relies on a non-linear combination of features extracted at different levels of a CNN. To validate our multilevel approach, we first evaluate the relative importance of features coming from each level. In the following, we define the importance of a feature as the extent to which a variation of the feature can affect the predicted map. Let us start by considering a linear model where different levels of a CNN are combined to obtain a pixel of the saliency map $\phi_i(\mathbf{x})$
\begin{equation}
\phi_i(\mathbf{x}) = w_i^T \mathbf{x} + \theta_i
\end{equation}
where $w_i$ and $\theta_i$ are the weight vector and the bias relative to pixel $i$, while $\mathbf{x}$ represents the activation coming from different levels of the feature extraction CNN, and $\phi_i(\mathbf{x})$ is the predicted saliency pixel. It is easy to see that the magnitude of elements in $w_i$ defines the importance of the corresponding features. 
In the extreme case of a pixel of the feature map which is always multiplied by 0, it is straightforward to see that part of the feature map is ignored by the model, and has therefore no importance, while a pixel with high absolute values in $w_i$ will have a considerable effect on the predicted saliency pixel.

In our model, $\phi_i(\cdot)$ is a highly non-linear function of the input, due to the presence of the encoding network and of the prior, thus the above reasoning is not directly applicable. Instead, given an image $\mathbf{x}_j$, we can approximate $\phi_i(\mathbf{x}_j)$ in the neighborhood of $\mathbf{x}_j$ as follows
\begin{equation}
\phi_i(\mathbf{x}_j) \approx \nabla \phi_i(\mathbf{x}_j)^T \mathbf{x} + \theta
\label{eq:approx}
\end{equation}
An intuitive explanation of this approximation is that the magnitude of the partial derivatives indicates which features need to be changed to affect the model. Also notice that Eq.~\ref{eq:approx} is equivalent to a first order Taylor expansion.

To get an estimation of the importance of each pixel in an activation map regardless of the choice of $\mathbf{x}_j$, we can average the element-wise absolute values of the gradient computed in the neighborhood of each validation sample.

\begin{equation}
w_i = \frac{1}{N}\sum_{j=1}^N \left[ \left| \frac{\partial \phi_i}{\partial x^1}(\mathbf{x}_j) \right|, \left| \frac{\partial \phi_i}{\partial x^2}(\mathbf{x}_j) \right|, \cdots, \left| \frac{\partial \phi_i}{\partial x^d}(\mathbf{x}_j) \right| \right]
\label{eq:abs}
\end{equation}
where $d$ is the dimensionality of $\mathbf{x}_j$. Then, to get the relative importance of each activation map, we average the values of $w_i$ corresponding to that map, and $L_1$ normalize the resulting importances.

To get an estimate of the importance of feature maps extracted from the CNN, we should compute $\phi_i(\mathbf{x}_j)$ for every test image $j$ and for every saliency pixel $i$. To reduce the amount of required computation, instead of computing the gradient of each saliency pixel, we compute the gradient of the mean and variance of the output saliency map, in the neighborhood of each test sample. Applying Eq.~\ref{eq:abs}, we get an indication of the contribution of each feature pixel to the mean and variance of the predicted map. 

\begin{figure}[b]
\centering
\subfigure[Contribution to mean] {
	\includegraphics[width=0.4\columnwidth]{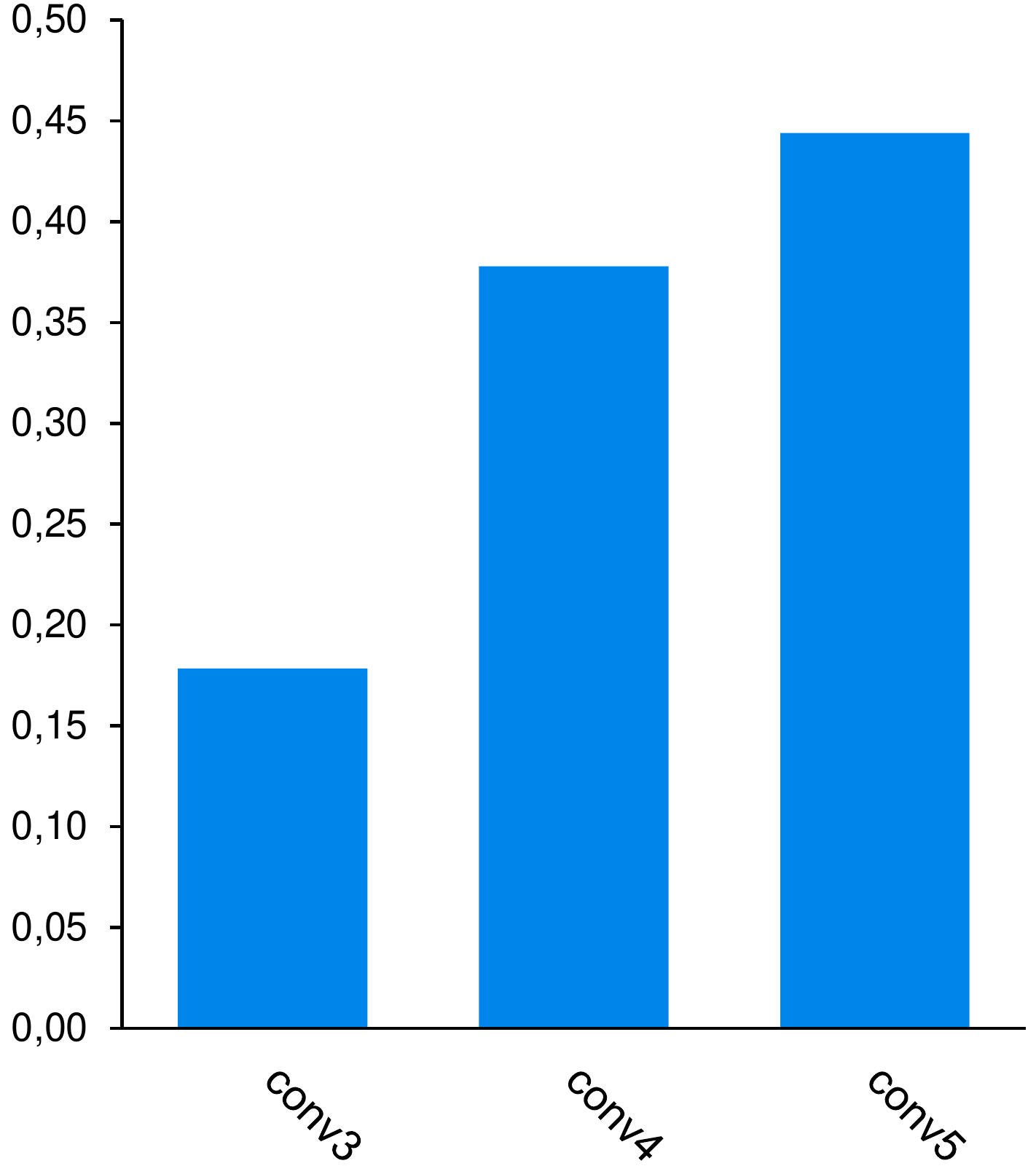}
}
\subfigure[Contribution to variance] {
	\includegraphics[width=0.4\columnwidth]{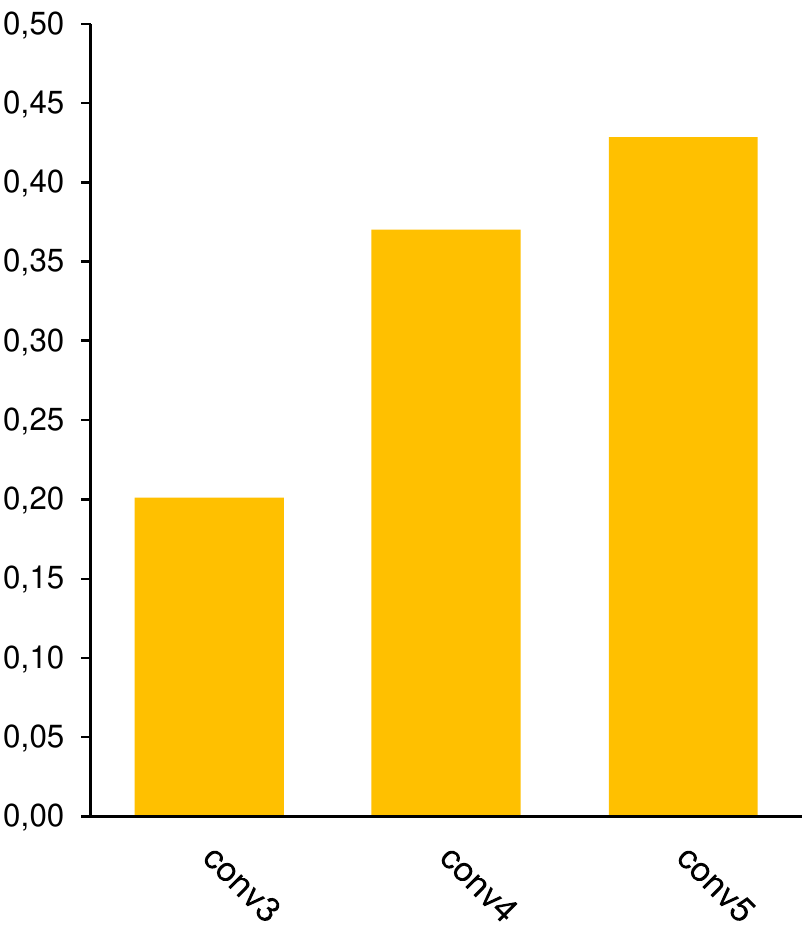}
}
\caption{Contribution of features extracted from \texttt{conv3}, \texttt{conv4} and \texttt{conv5} to prediction mean and variance.}
\label{fig:importance}
\end{figure}

Figure~\ref{fig:importance} reports the relative importance of activation maps coming from \texttt{conv3}, \texttt{conv4} and \texttt{conv5} on the model trained on SALICON. It is easy to notice that all features give a valuable contribution to the final result, and that while high level features are still the most relevant ones, medium level features have a considerable role in the prediction of the saliency map. This confirms our strategy to incorporate activations coming from different levels.

\subsection{Comparison with state of the art}
We evaluate our model on the SALICON dataset and on the MIT300 benchmark. In the first case, the network is trained on training images from the SALICON dataset, in the latter after training on SALICON we finetune on the MIT1003 dataset, as suggested by the MIT300 benchmark organizers. 
Images from all datasets were zero-padded to fit a $4:3$ aspect ratio, and then resized to $640 \times 480$.

Table~\ref{tab:salicon} compares the performance of our model on SALICON in terms of CC, AUC shuffled and AUC Judd. As it can be observed, our solution outperforms all competitors by a margin of 12\% according to CC metric, 4\% and 1\% according to AUC shuffled and AUC Judd.

For reference, the proposed solution is also evaluated on the MIT300 saliency benchmark which contains results of almost 60 different methods. Table~\ref{tab:mit300} presents a comparison between our approach and top performers in this benchmark. Our method outperforms the vast majority of the approaches in the leaderboard of the benchmark, and achieves competitive results when compared to the top ranked approaches.

\begin{figure*}[htbp]
\centering
\setlength\tabcolsep{1.2pt}
\begin{tabular}{cccccccccccccc}
\footnotesize{Image} & \footnotesize{GT} & \footnotesize{Our} & \tiny{Deep~\cite{cPan}} & \tiny{Shallow~\cite{cPan}} & \footnotesize{\cite{riche2013rare2012}} & \footnotesize{\cite{harel2006graph}} & \footnotesize{Image} & \footnotesize{GT} & \footnotesize{Our} & \footnotesize{\cite{liu2015predicting}} & \footnotesize{\cite{vig2014large}} & \footnotesize{\cite{riche2013rare2012}} & \footnotesize{\cite{harel2006graph}} \\
\subfigure{\includegraphics[width=0.065\textwidth]{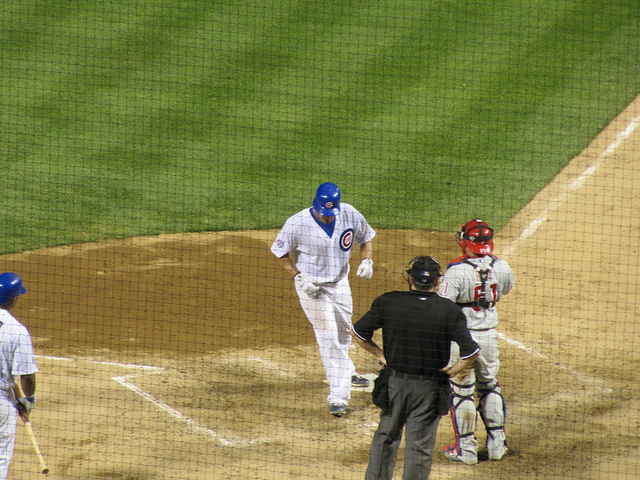}}&
\subfigure{\includegraphics[width=0.065\textwidth]{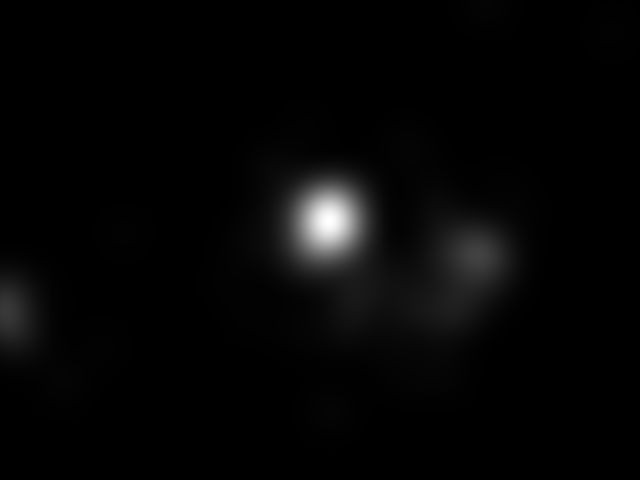}}&
\subfigure{\includegraphics[width=0.065\textwidth]{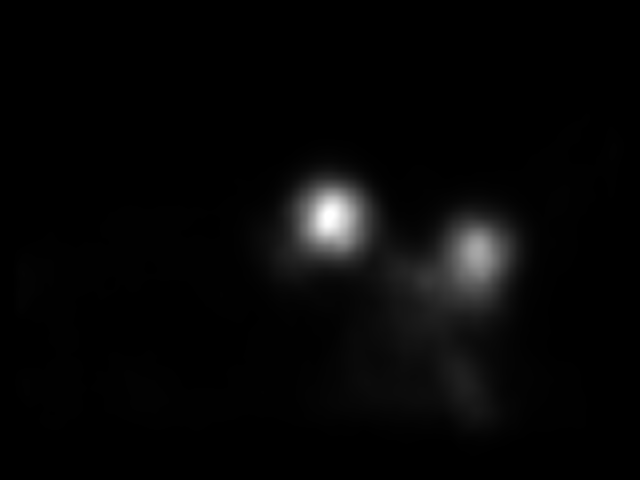}}&
\subfigure{\includegraphics[width=0.065\textwidth]{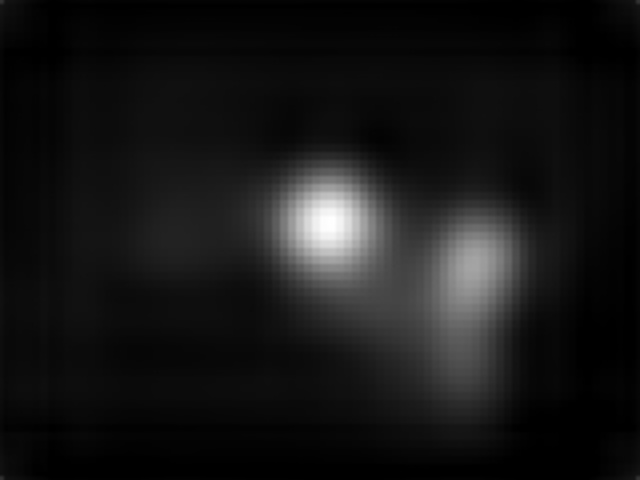}}&
\subfigure{\includegraphics[width=0.065\textwidth]{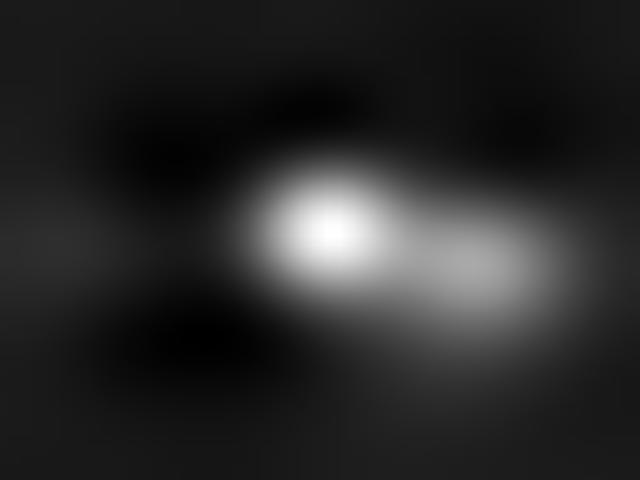}}&
\subfigure{\includegraphics[width=0.065\textwidth]{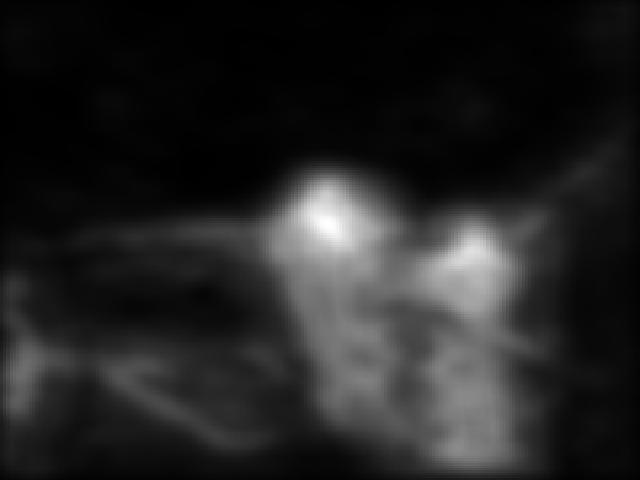}}&
\subfigure{\includegraphics[width=0.065\textwidth]{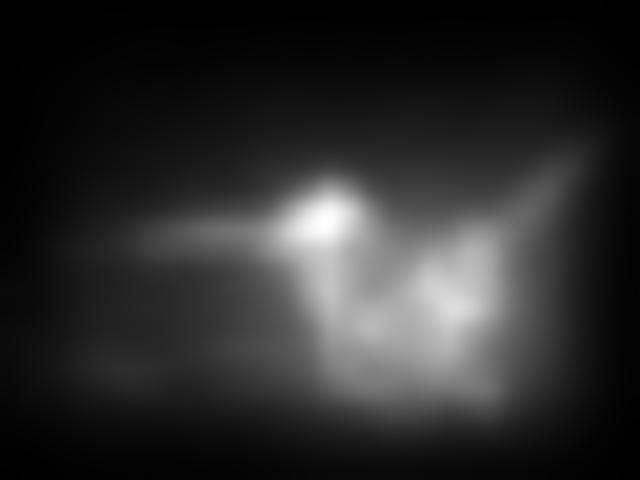}} &
\subfigure{\includegraphics[width=0.065\textwidth]{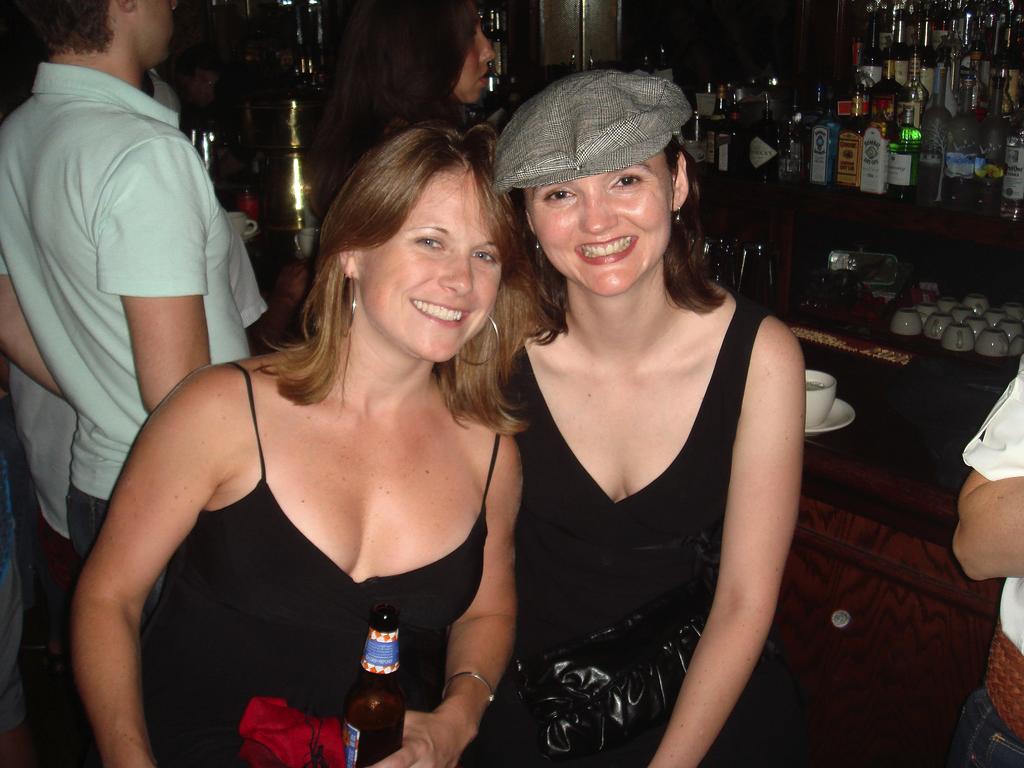}}&
\subfigure{\includegraphics[width=0.065\textwidth]{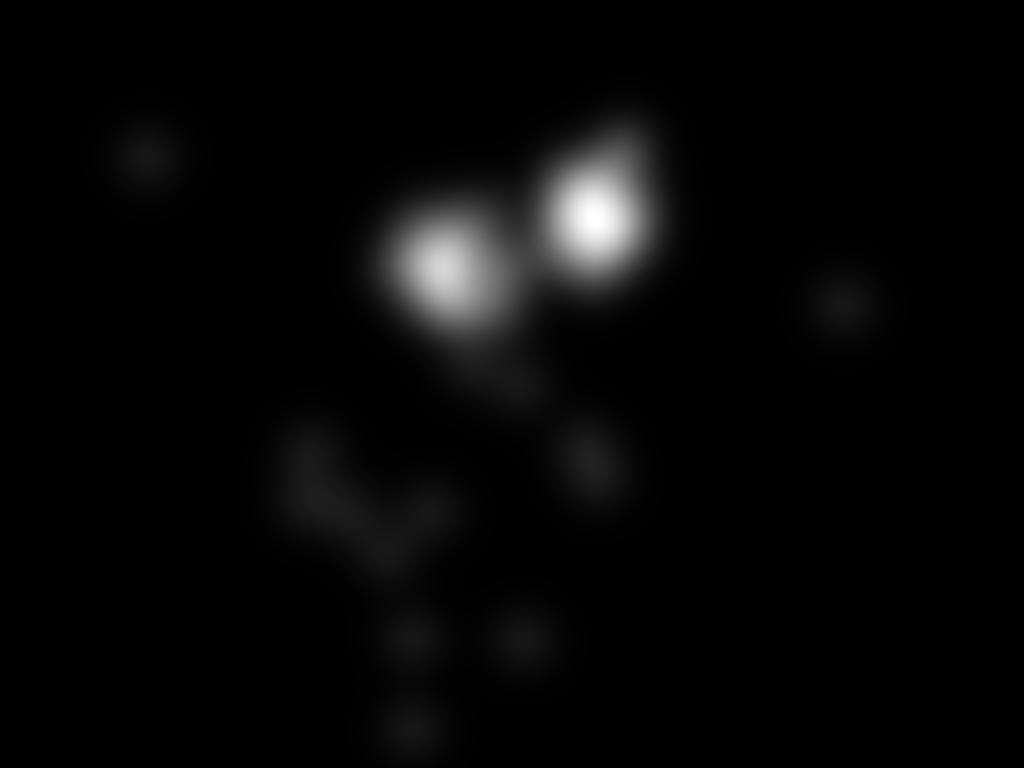}}&
\subfigure{\includegraphics[width=0.065\textwidth]{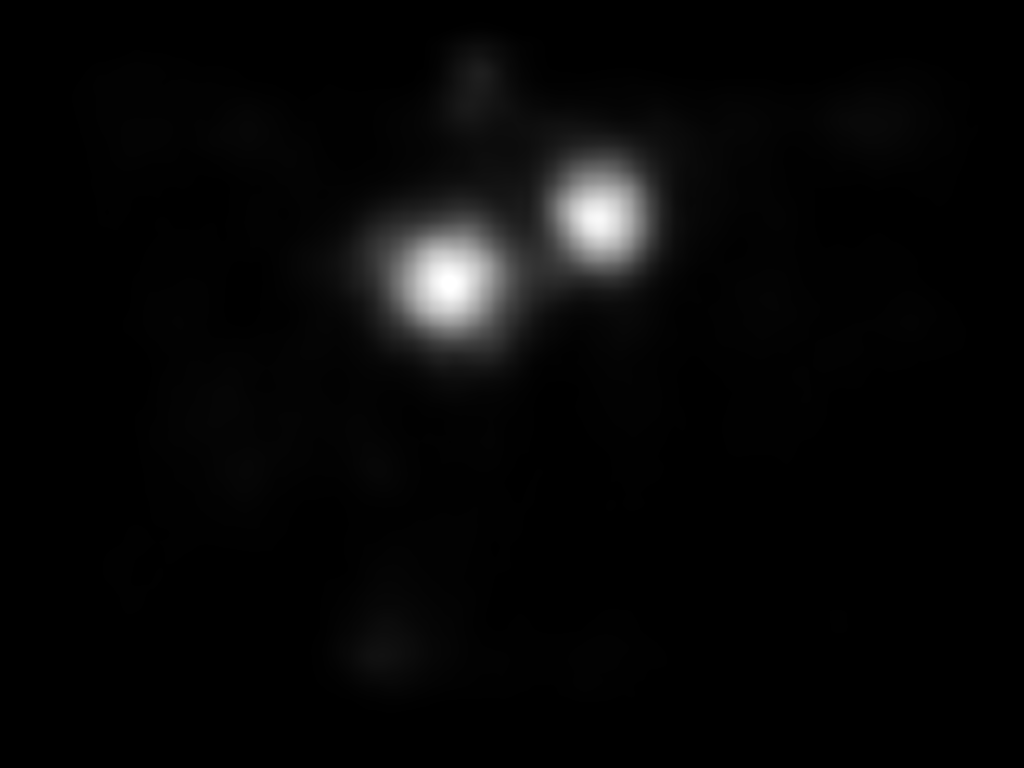}}&
\subfigure{\includegraphics[width=0.065\textwidth]{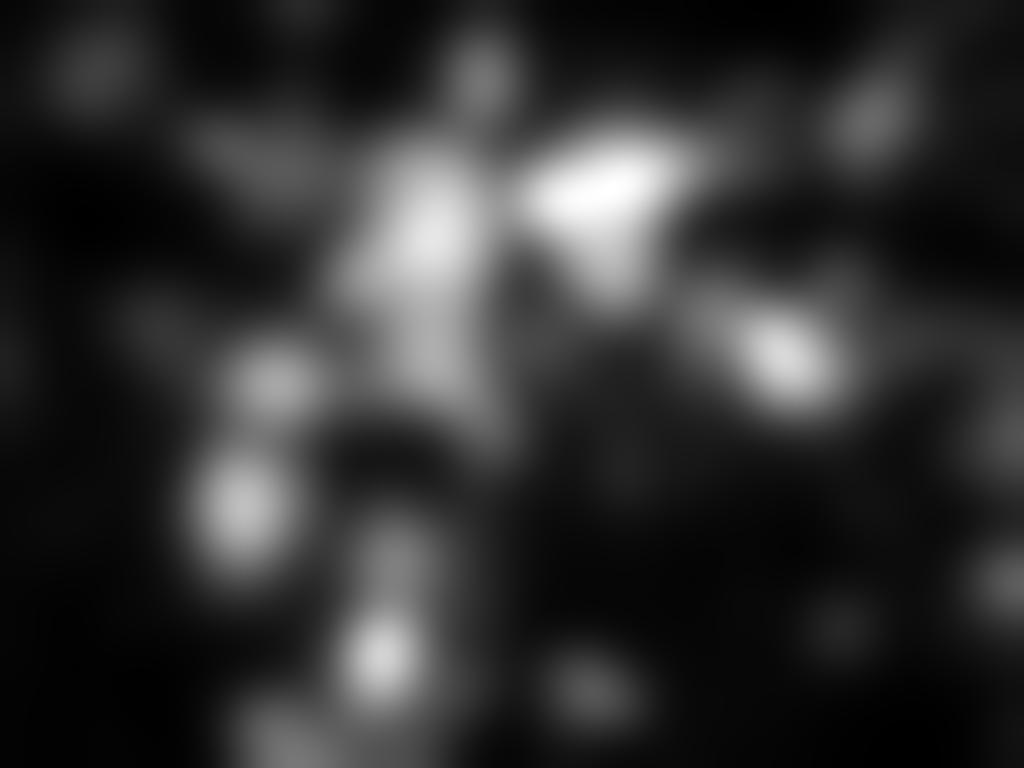}}&
\subfigure{\includegraphics[width=0.065\textwidth]{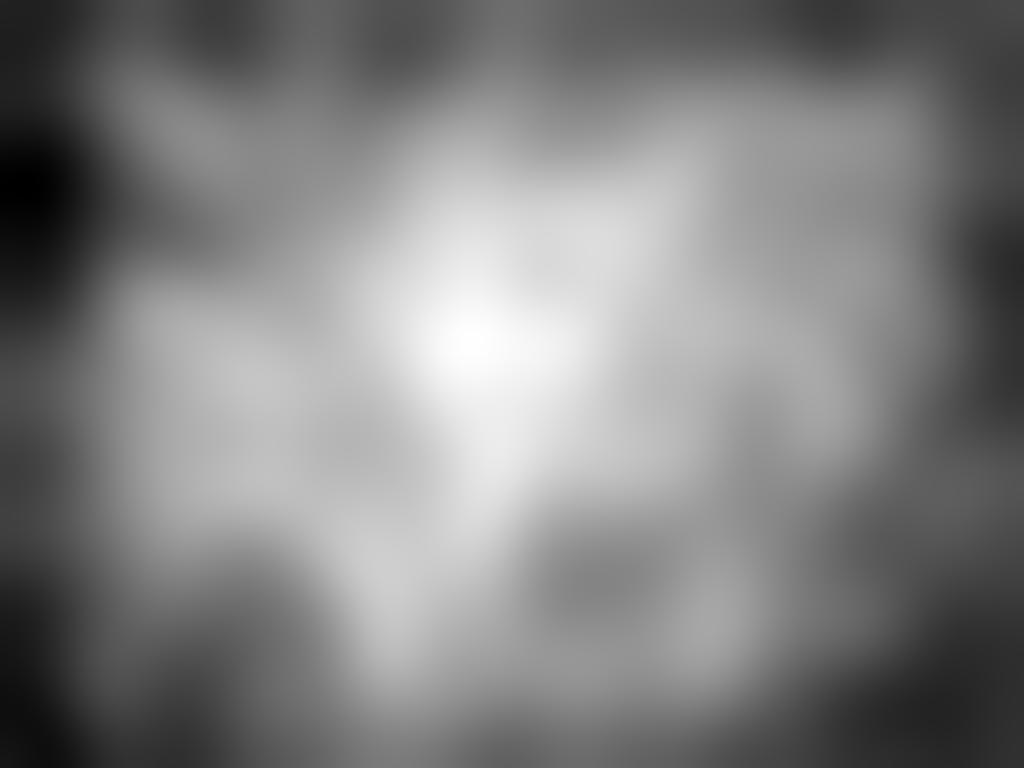}}&
\subfigure{\includegraphics[width=0.065\textwidth]{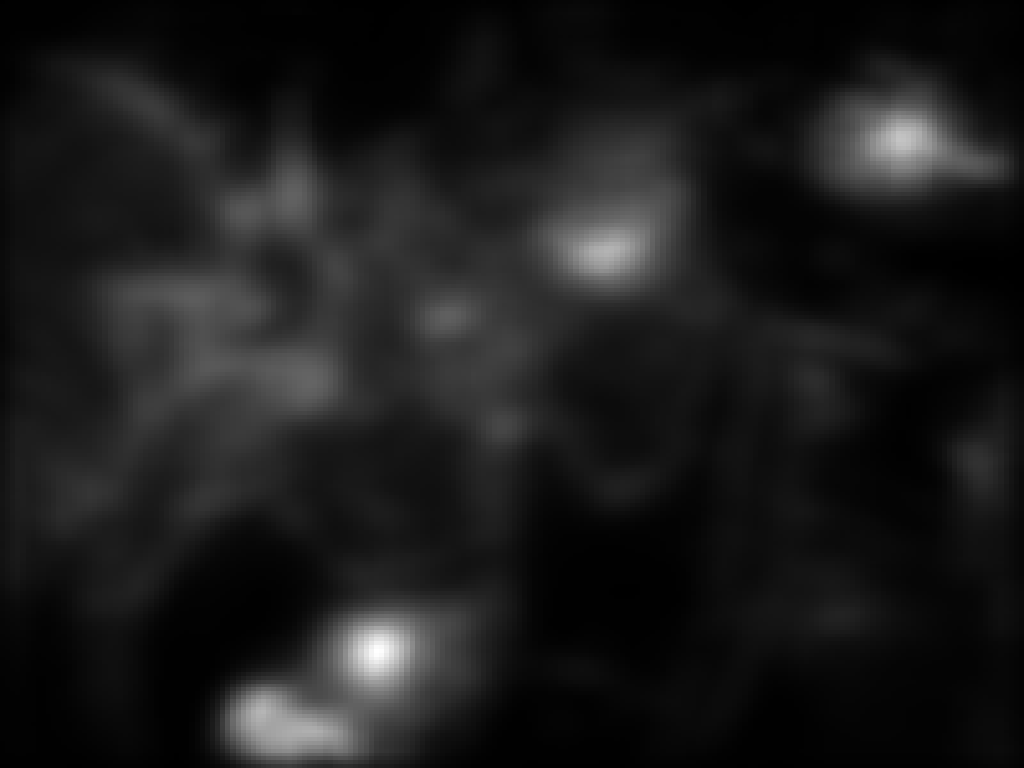}}&
\subfigure{\includegraphics[width=0.065\textwidth]{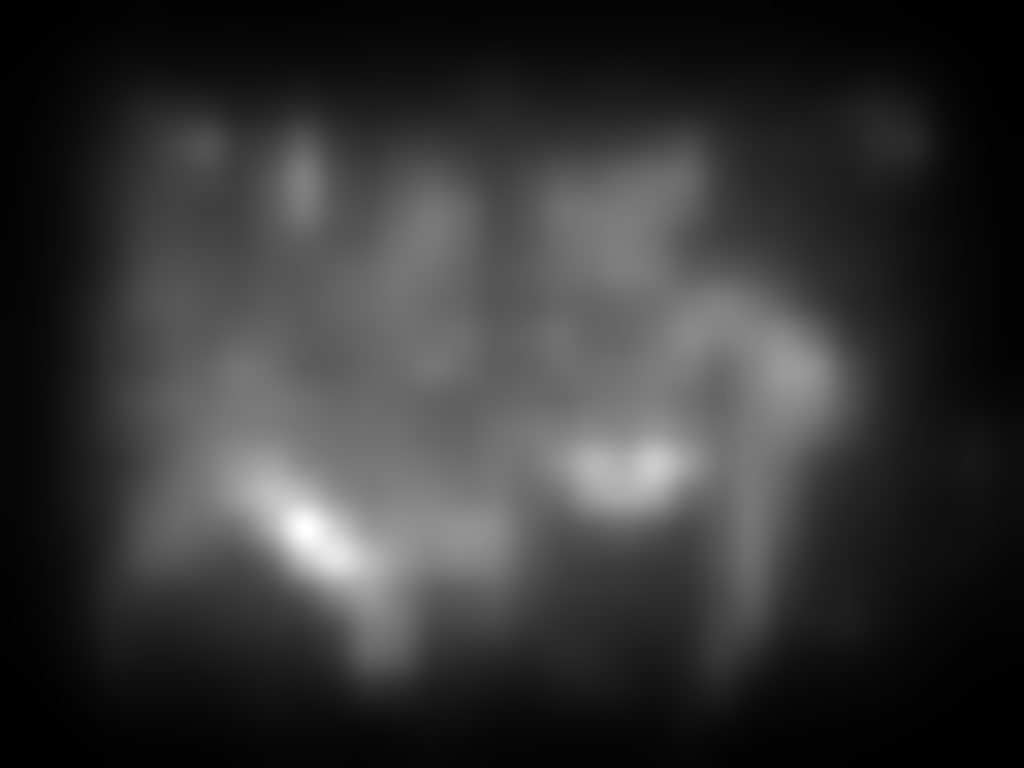}}\\

\subfigure{\includegraphics[width=0.065\textwidth]{COCOval2014000000001532.jpg}}&
\subfigure{\includegraphics[width=0.065\textwidth]{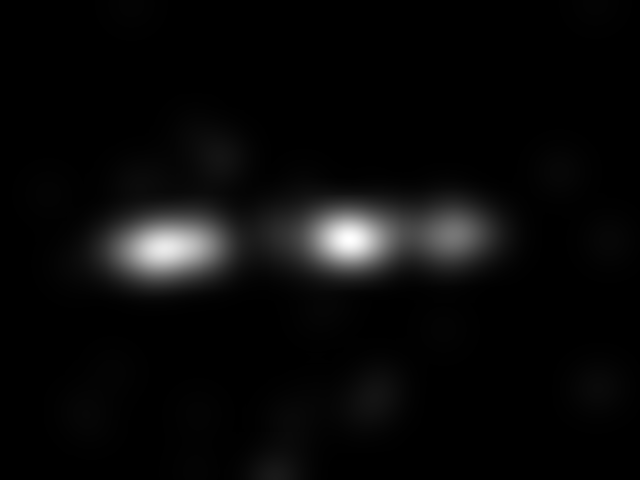}}&
\subfigure{\includegraphics[width=0.065\textwidth]{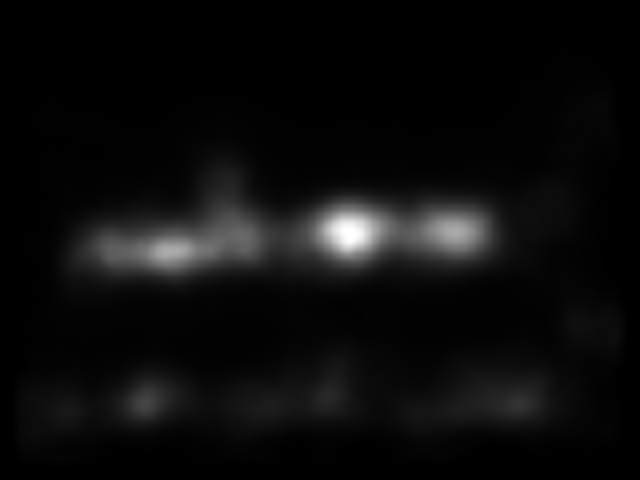}}&
\subfigure{\includegraphics[width=0.065\textwidth]{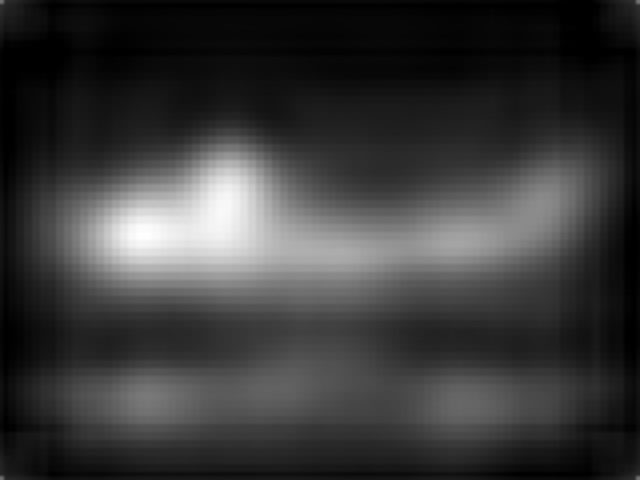}}&
\subfigure{\includegraphics[width=0.065\textwidth]{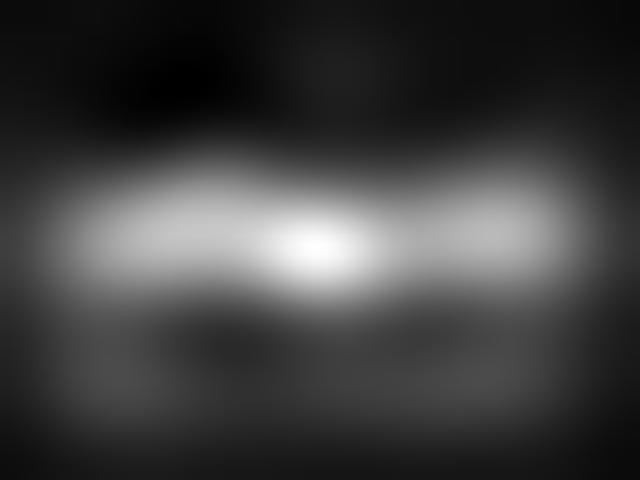}}&
\subfigure{\includegraphics[width=0.065\textwidth]{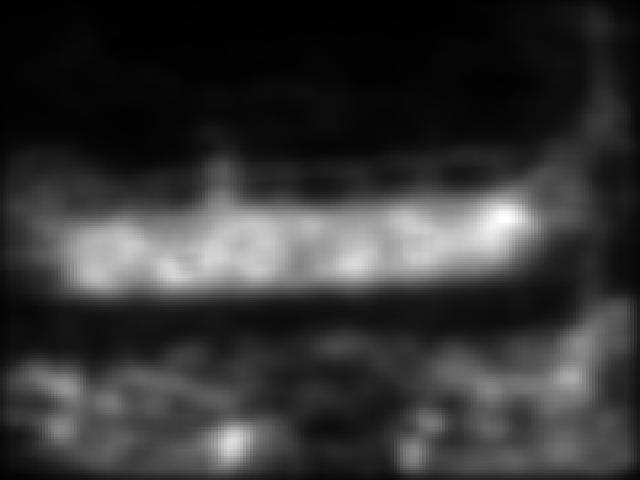}}&
\subfigure{\includegraphics[width=0.065\textwidth]{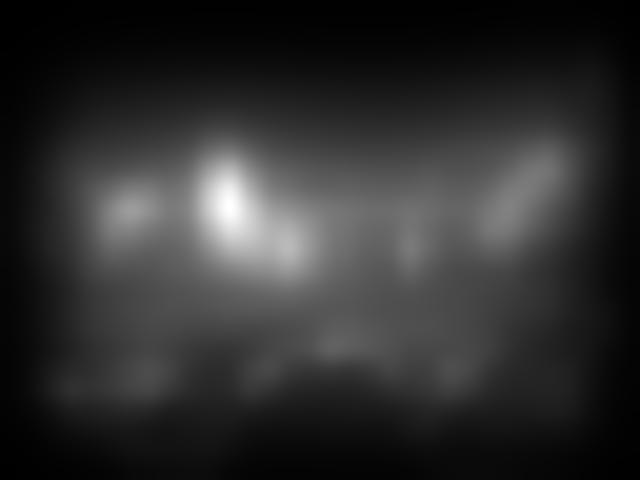}}&

\subfigure{\includegraphics[width=0.065\textwidth]{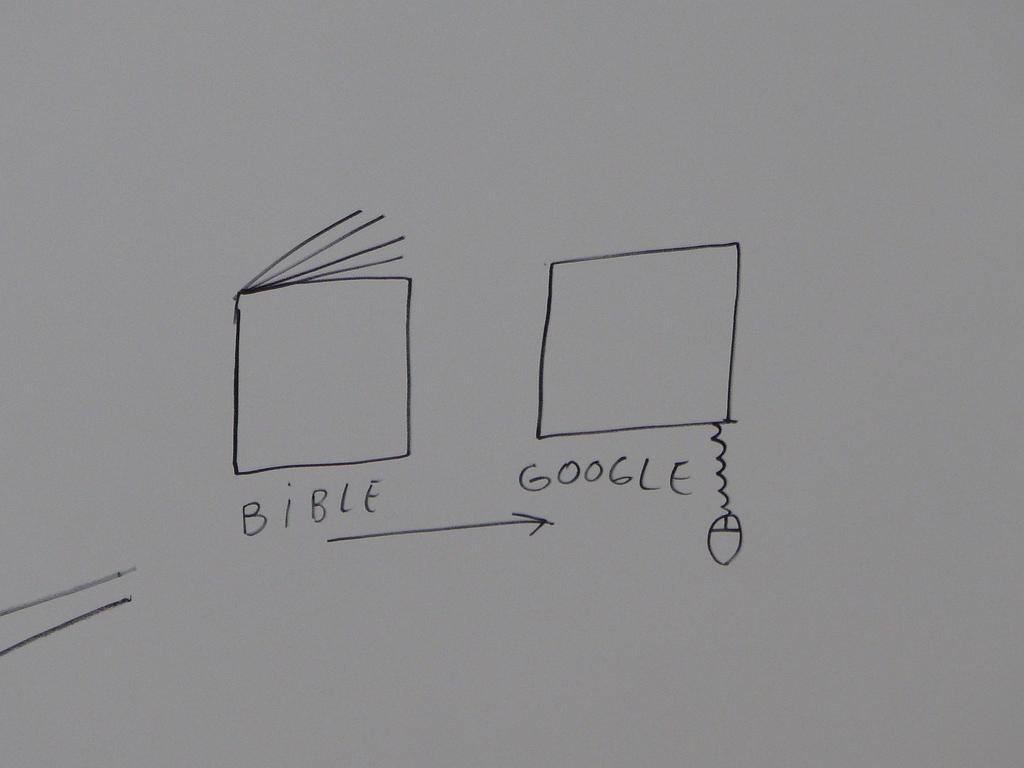}}&
\subfigure{\includegraphics[width=0.065\textwidth]{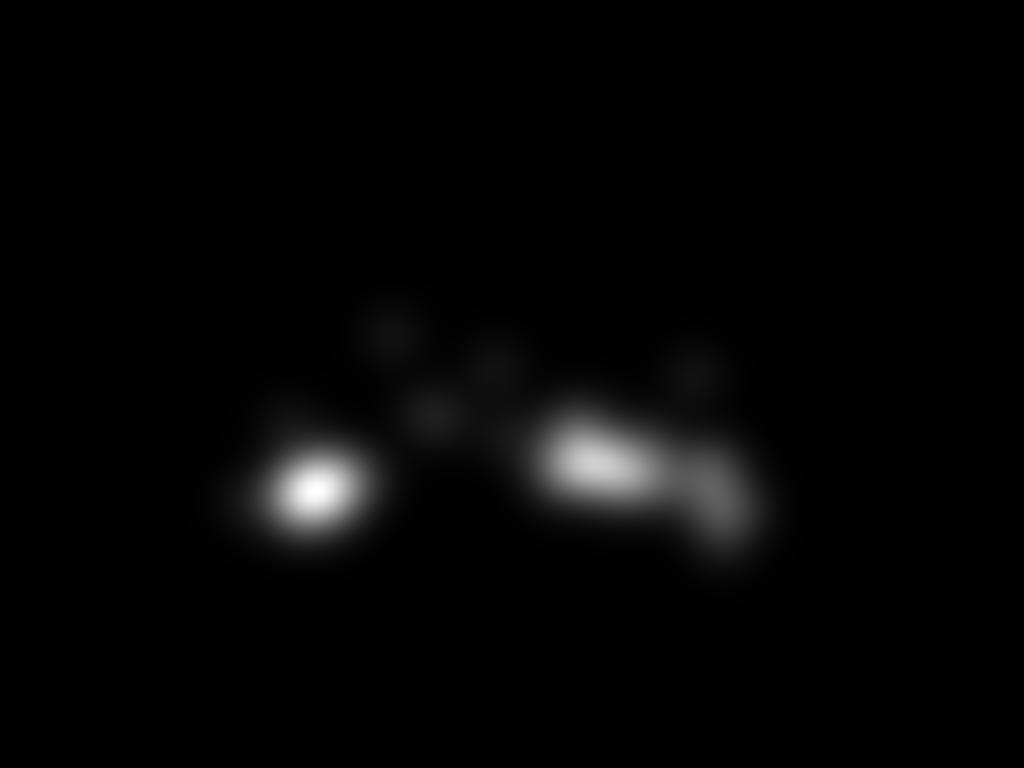}}&
\subfigure{\includegraphics[width=0.065\textwidth]{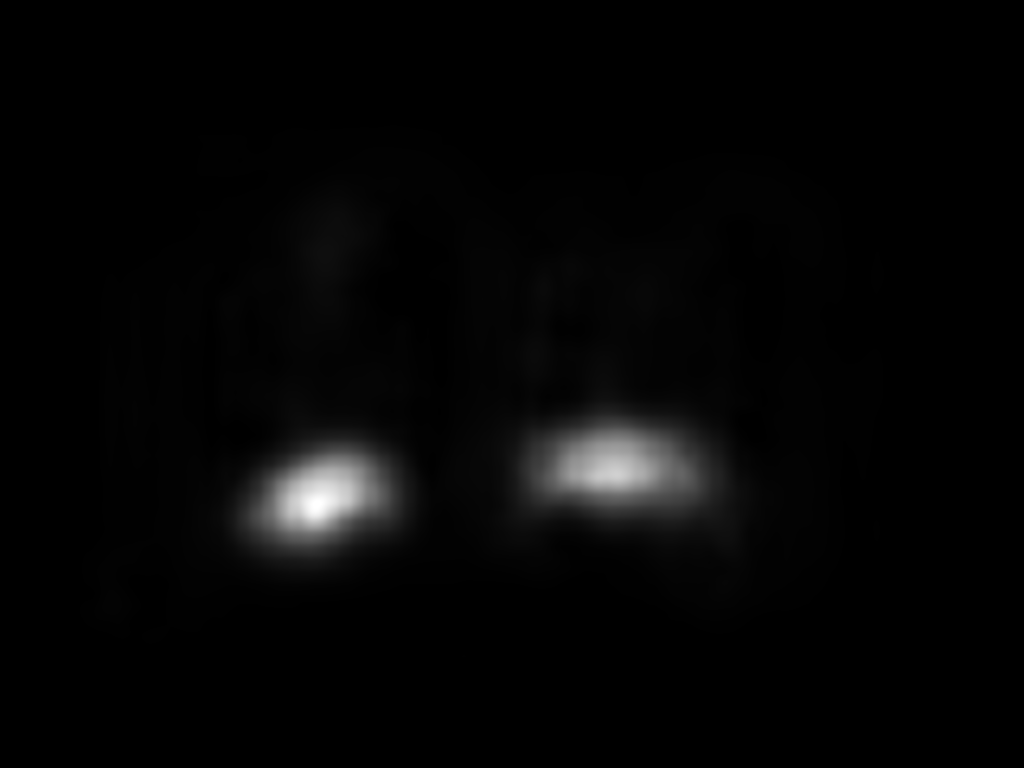}}&
\subfigure{\includegraphics[width=0.065\textwidth]{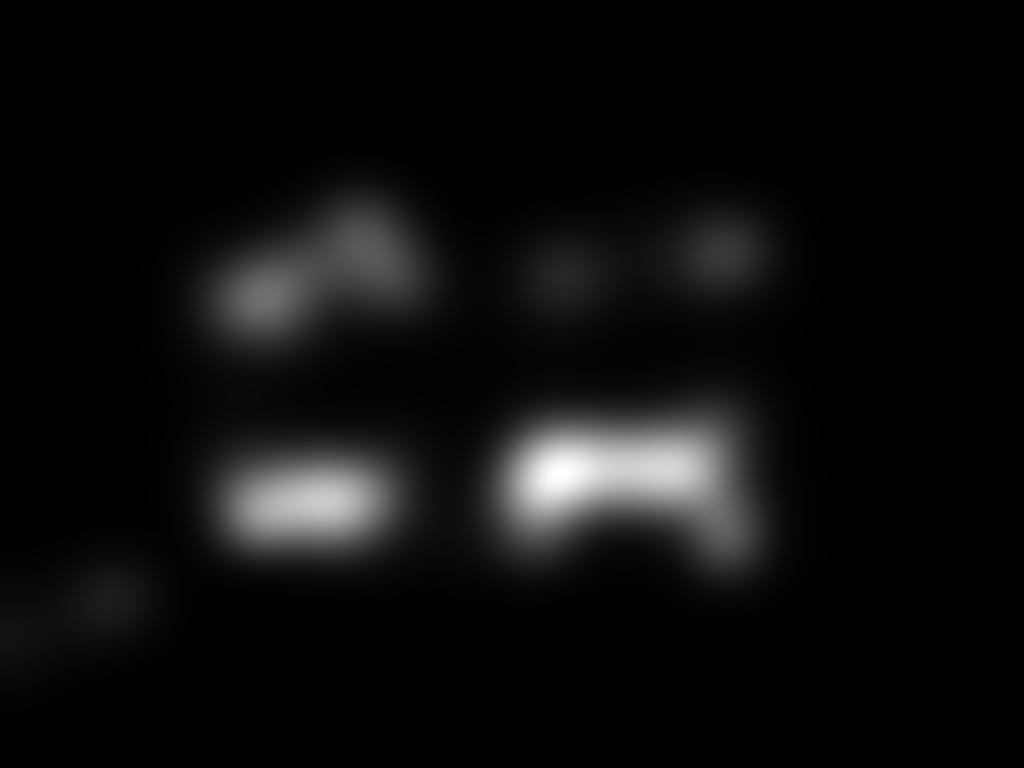}}&
\subfigure{\includegraphics[width=0.065\textwidth]{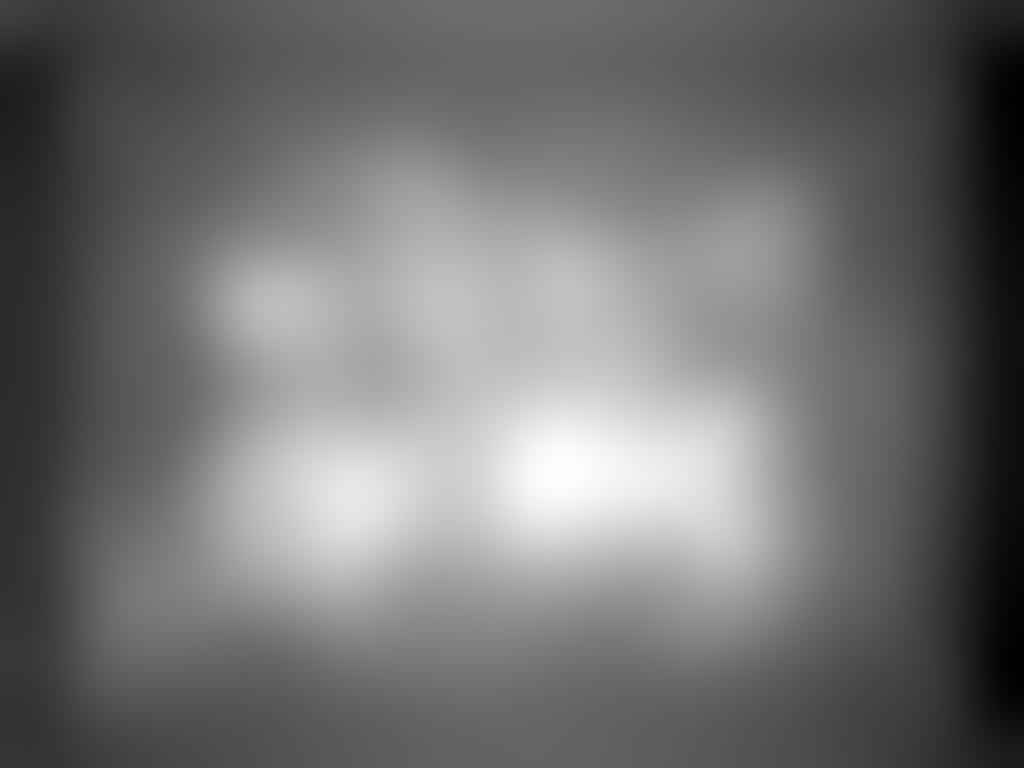}}&
\subfigure{\includegraphics[width=0.065\textwidth]{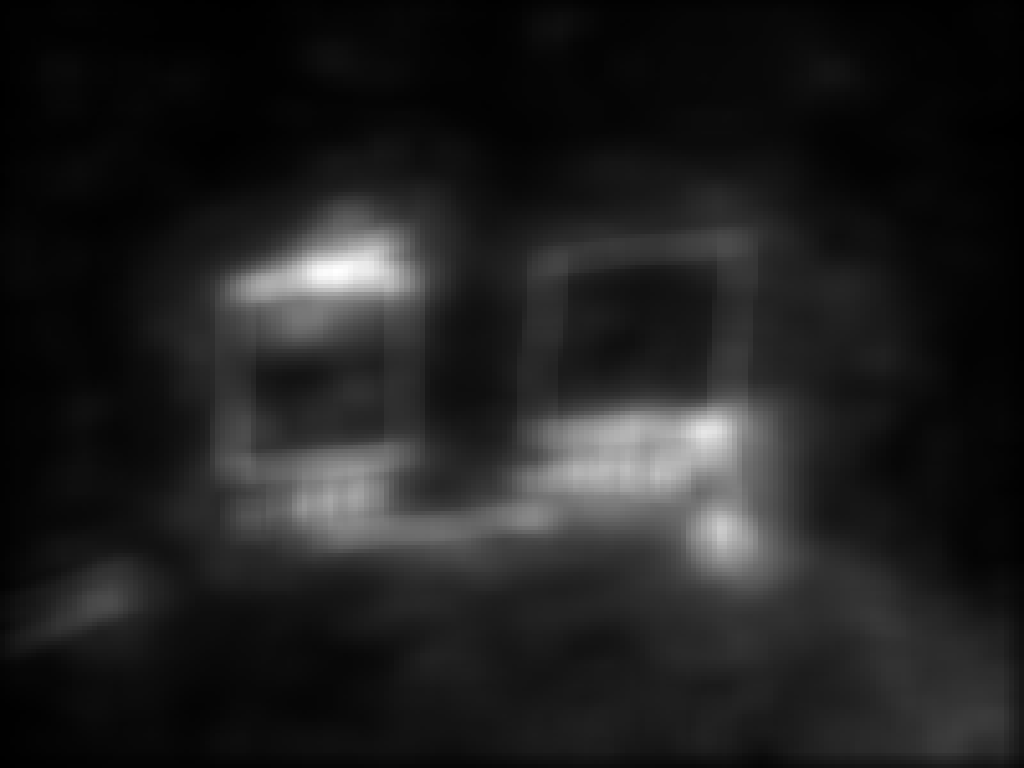}}&
\subfigure{\includegraphics[width=0.065\textwidth]{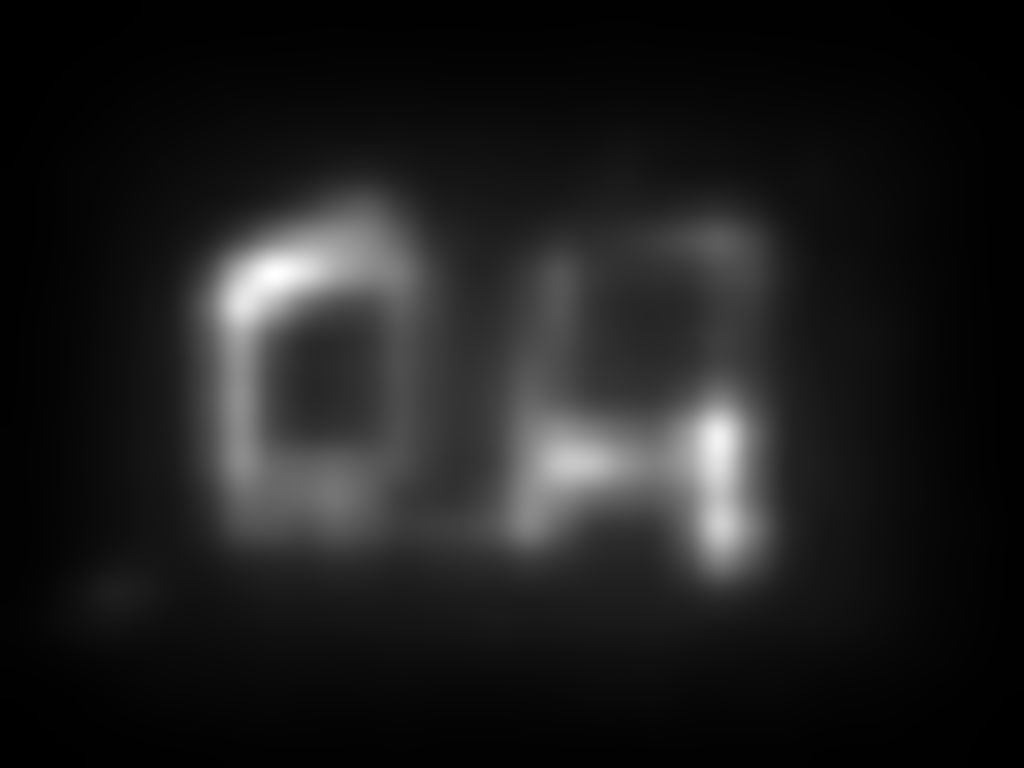}}\\

\subfigure{\includegraphics[width=0.065\textwidth]{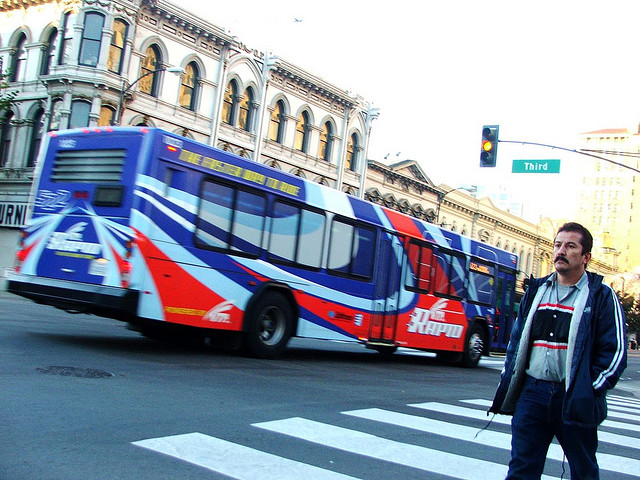}}&
\subfigure{\includegraphics[width=0.065\textwidth]{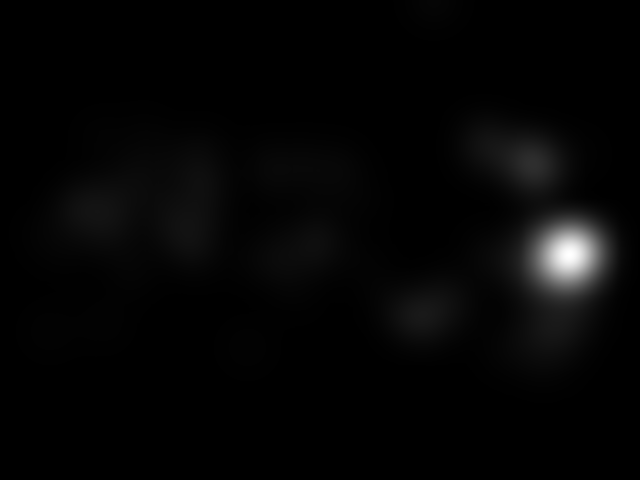}}&
\subfigure{\includegraphics[width=0.065\textwidth]{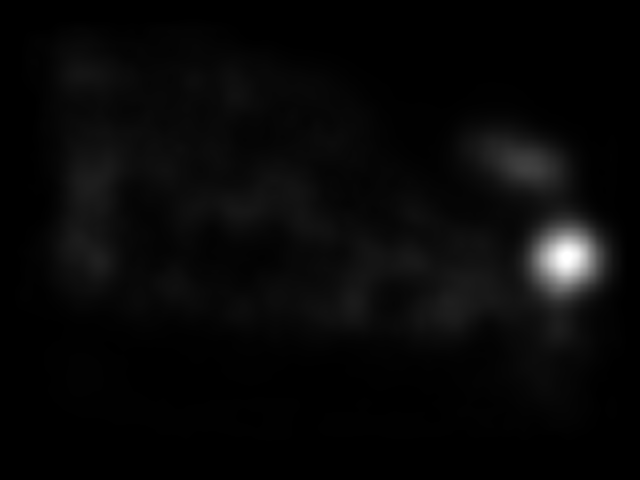}}&
\subfigure{\includegraphics[width=0.065\textwidth]{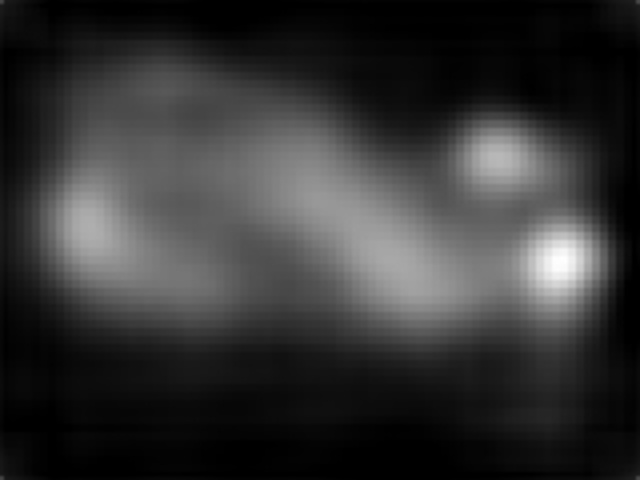}}&
\subfigure{\includegraphics[width=0.065\textwidth]{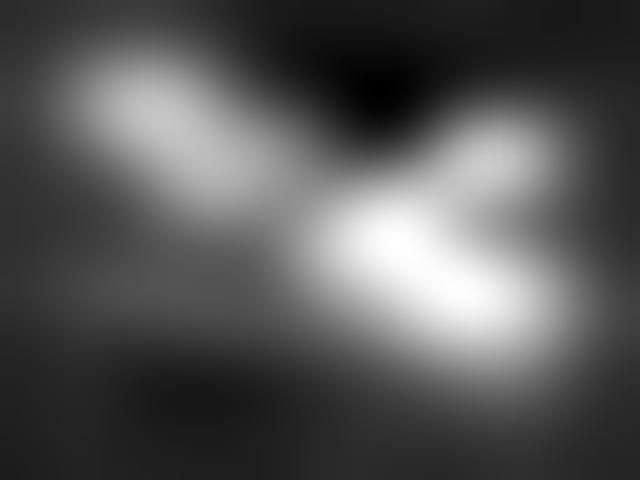}}&
\subfigure{\includegraphics[width=0.065\textwidth]{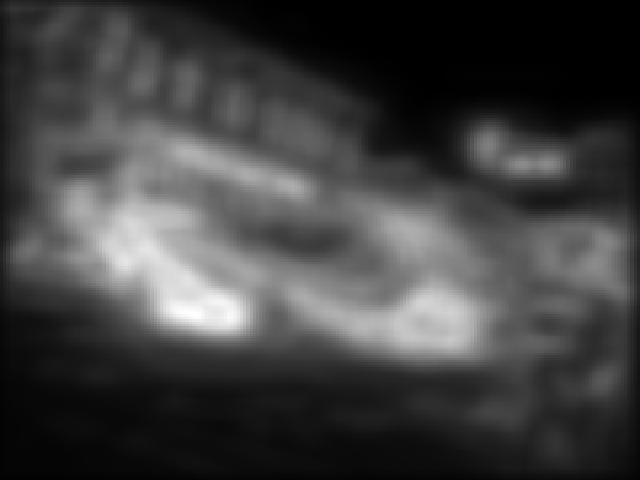}}&
\subfigure{\includegraphics[width=0.065\textwidth]{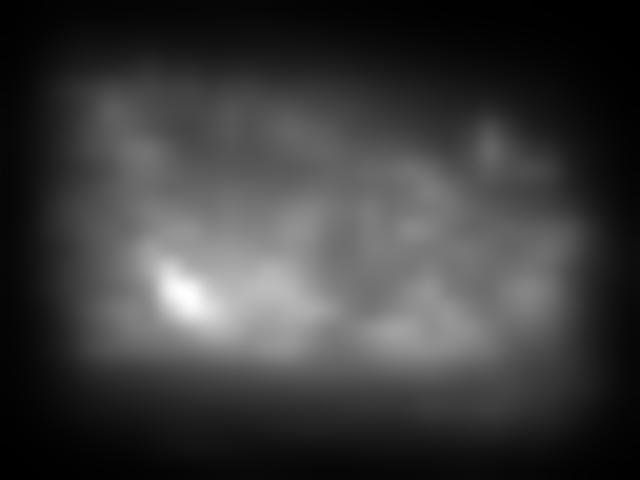}}&

\subfigure{\includegraphics[width=0.065\textwidth]{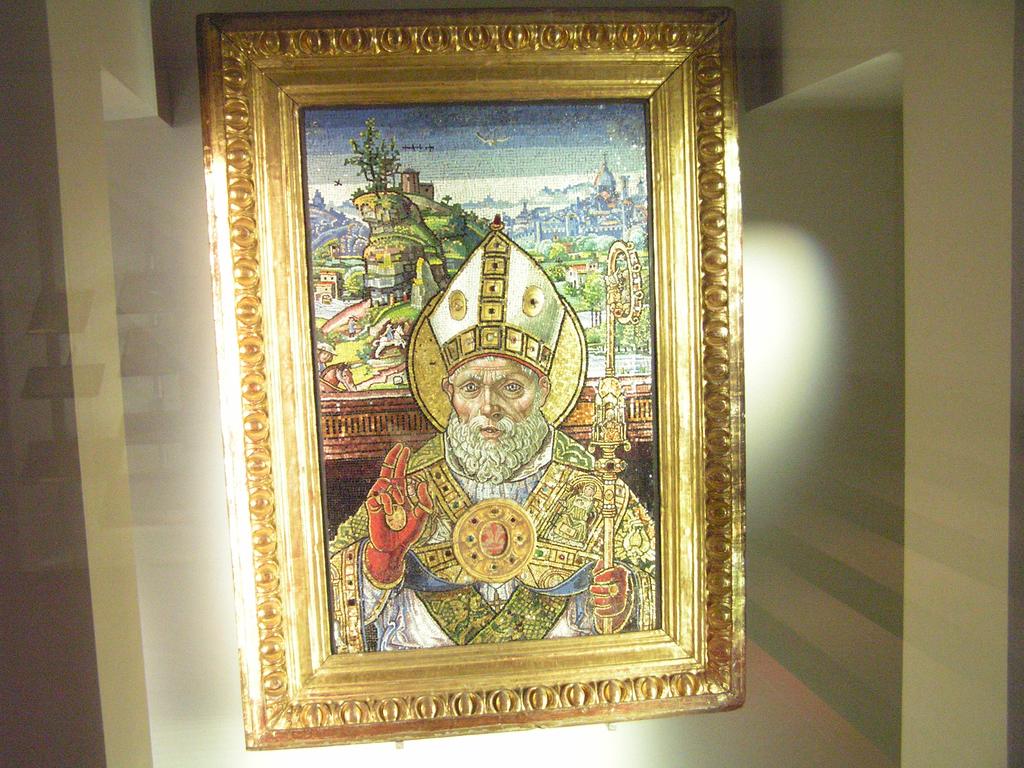}}&
\subfigure{\includegraphics[width=0.065\textwidth]{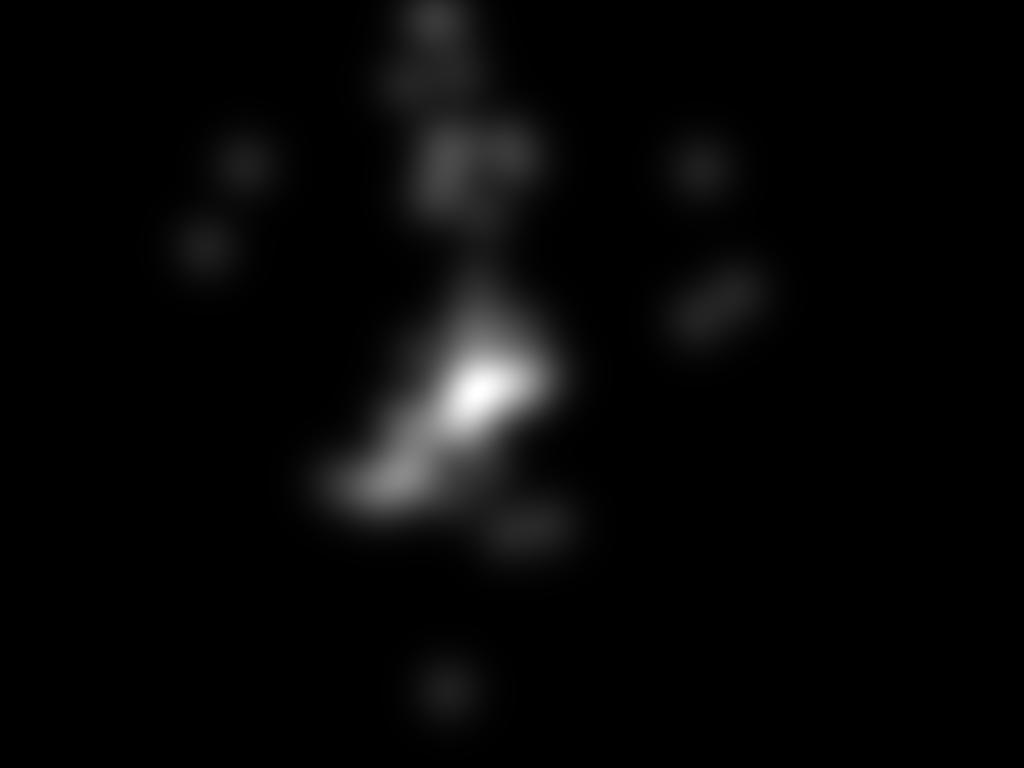}}&
\subfigure{\includegraphics[width=0.065\textwidth]{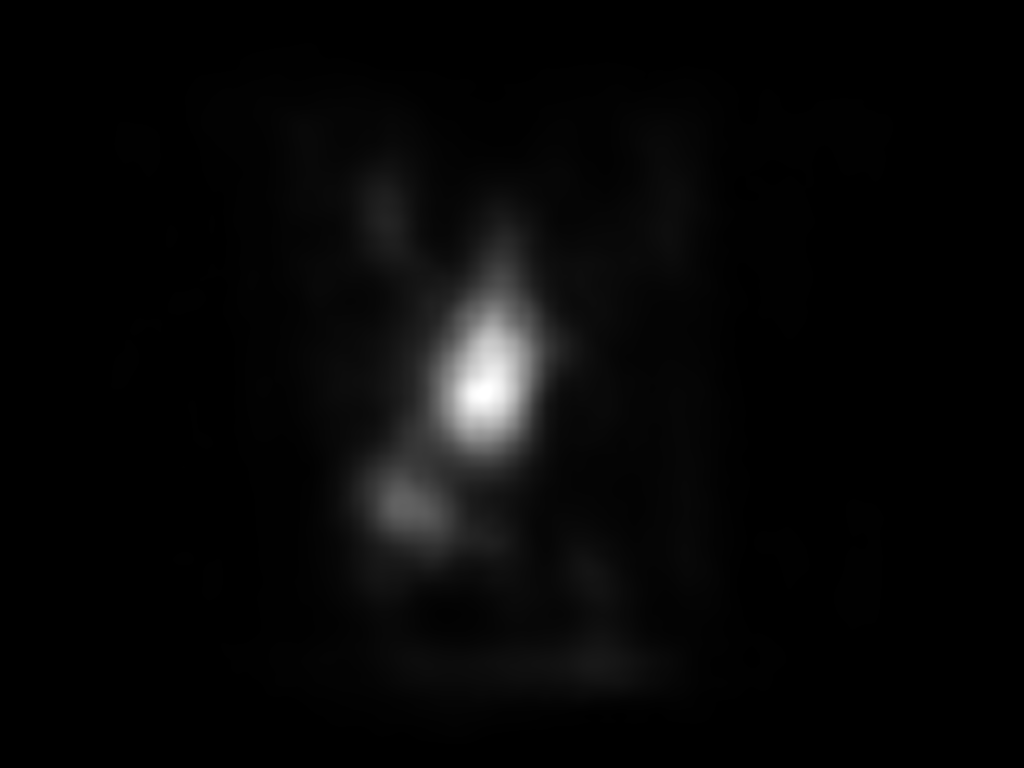}}&
\subfigure{\includegraphics[width=0.065\textwidth]{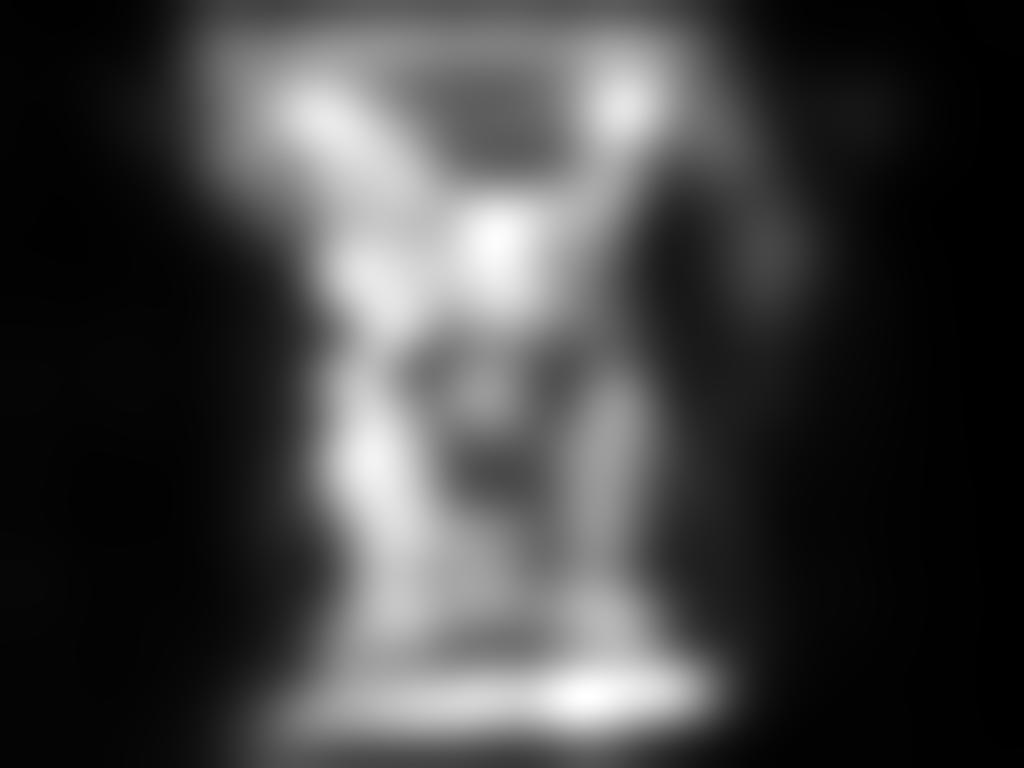}}&
\subfigure{\includegraphics[width=0.065\textwidth]{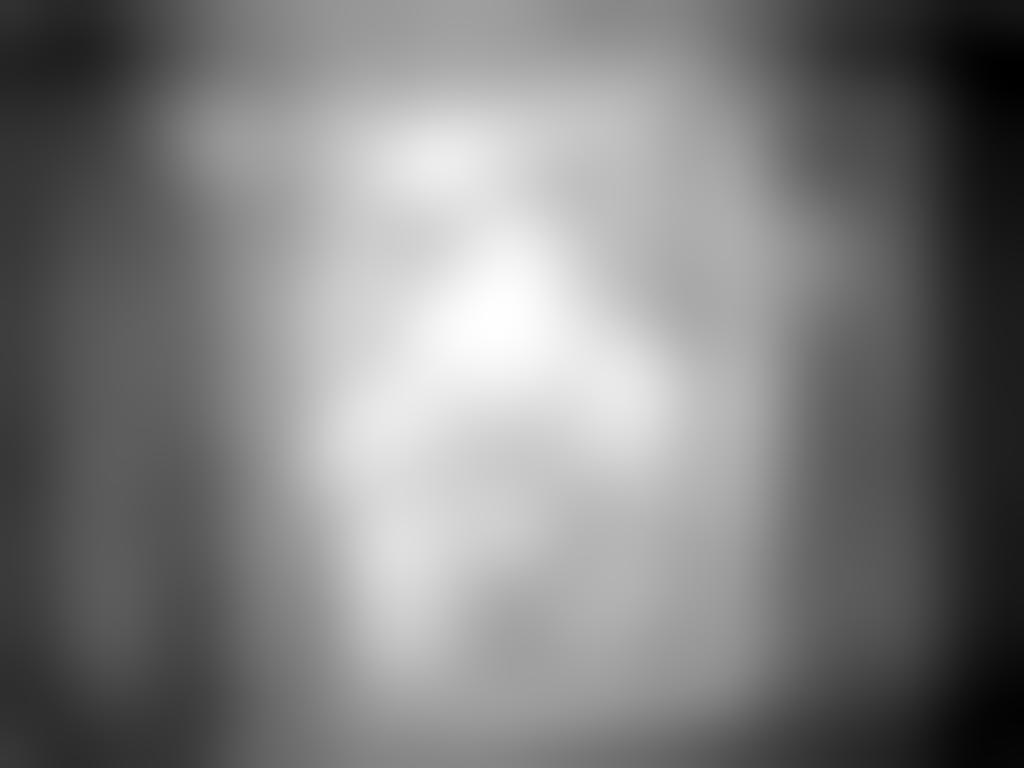}}&
\subfigure{\includegraphics[width=0.065\textwidth]{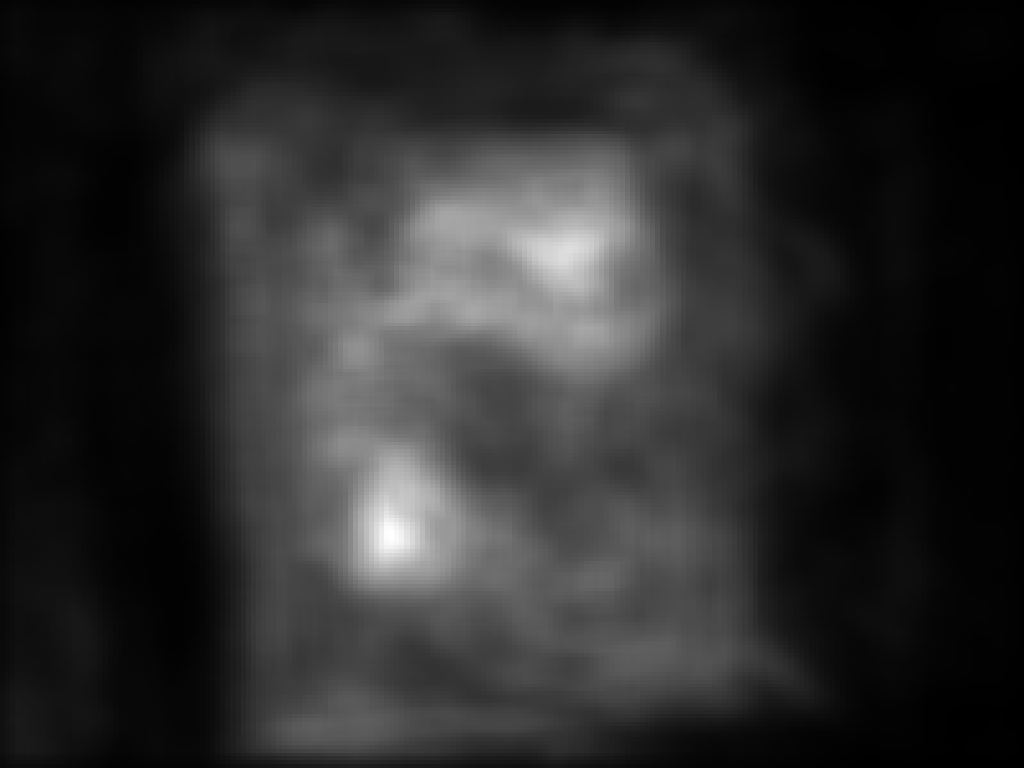}}&
\subfigure{\includegraphics[width=0.065\textwidth]{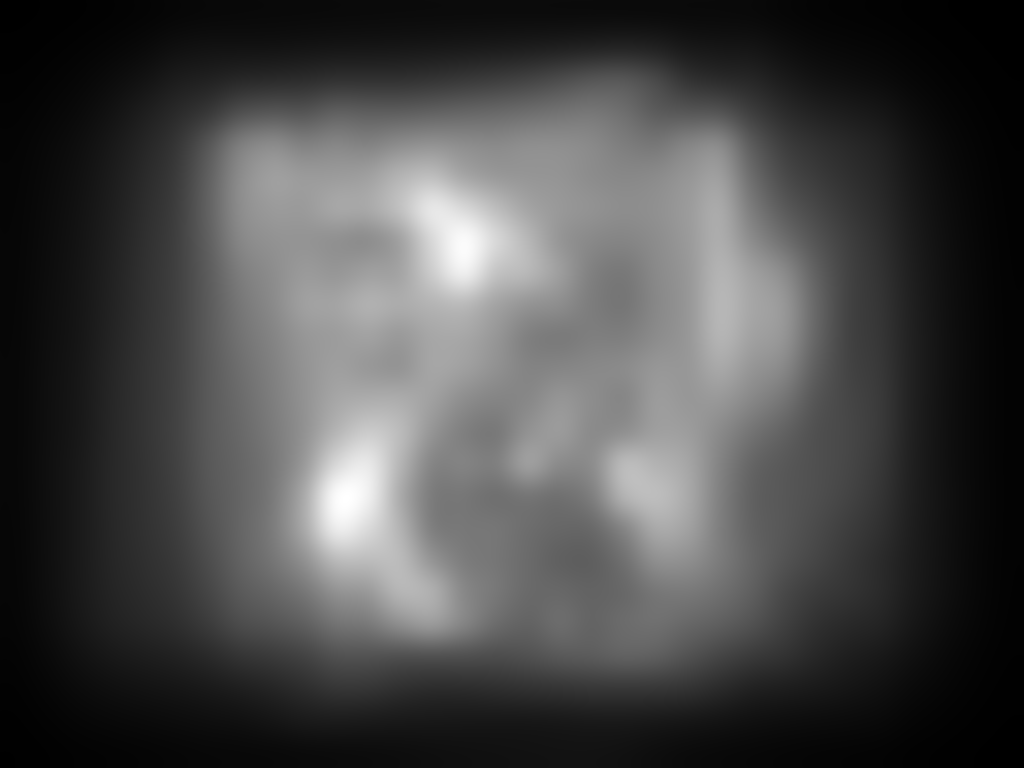}} \\

\subfigure{\includegraphics[width=0.065\textwidth]{COCOval2014000000001000.jpg}}&
\subfigure{\includegraphics[width=0.065\textwidth]{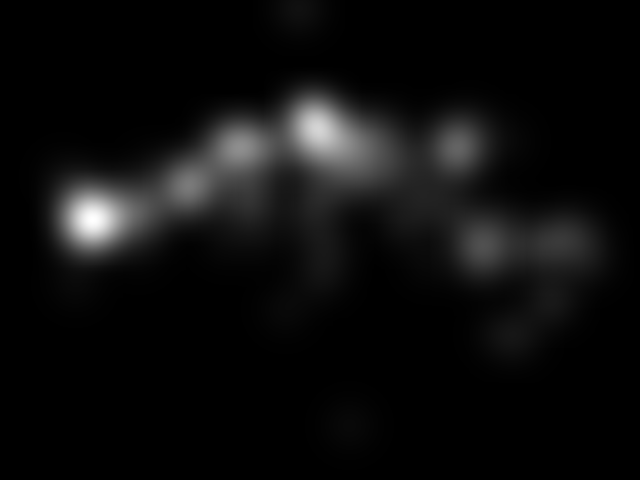}}&
\subfigure{\includegraphics[width=0.065\textwidth]{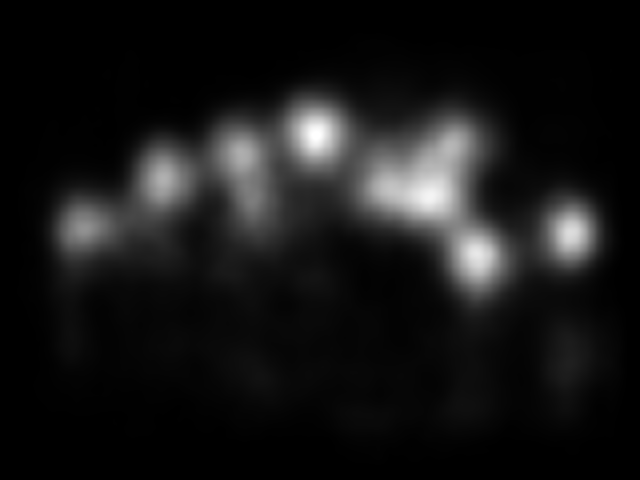}}&
\subfigure{\includegraphics[width=0.065\textwidth]{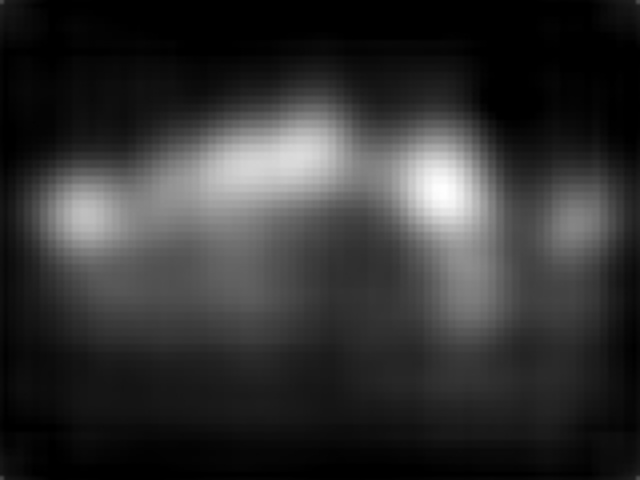}}&
\subfigure{\includegraphics[width=0.065\textwidth]{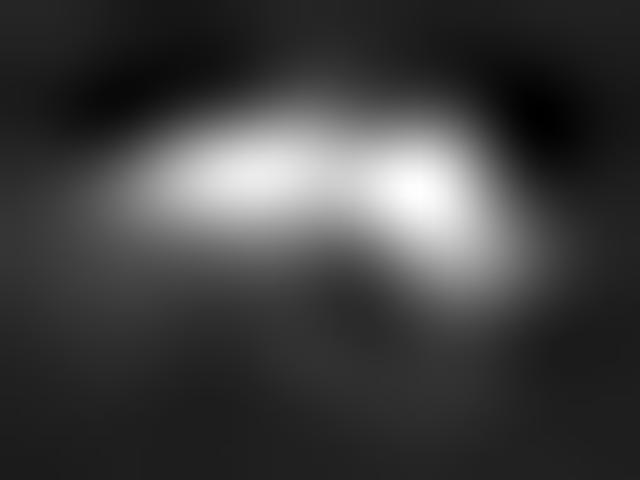}}&
\subfigure{\includegraphics[width=0.065\textwidth]{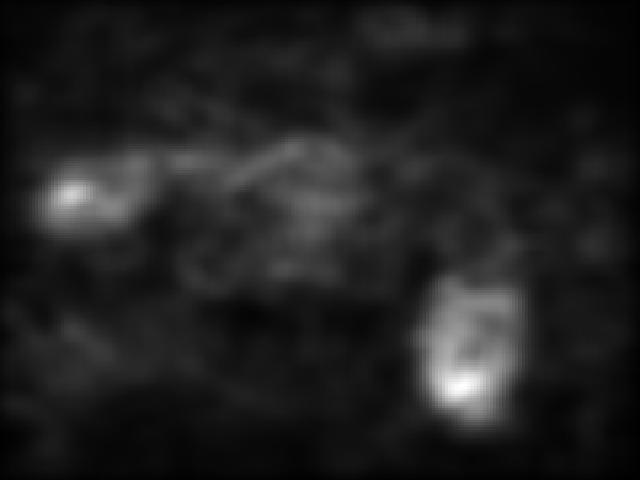}}&
\subfigure{\includegraphics[width=0.065\textwidth]{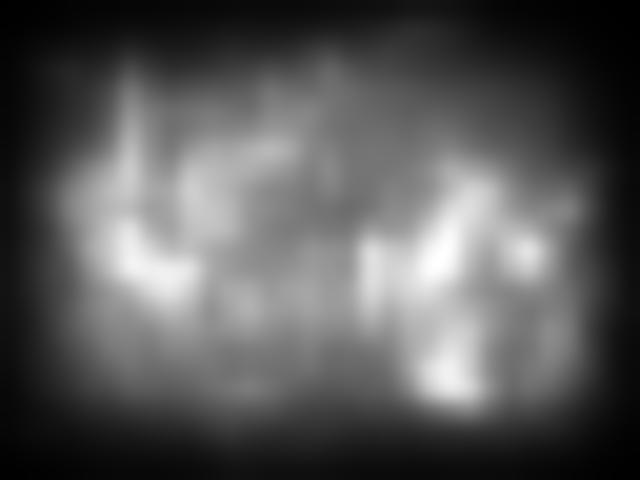}}&

\subfigure{\includegraphics[width=0.065\textwidth]{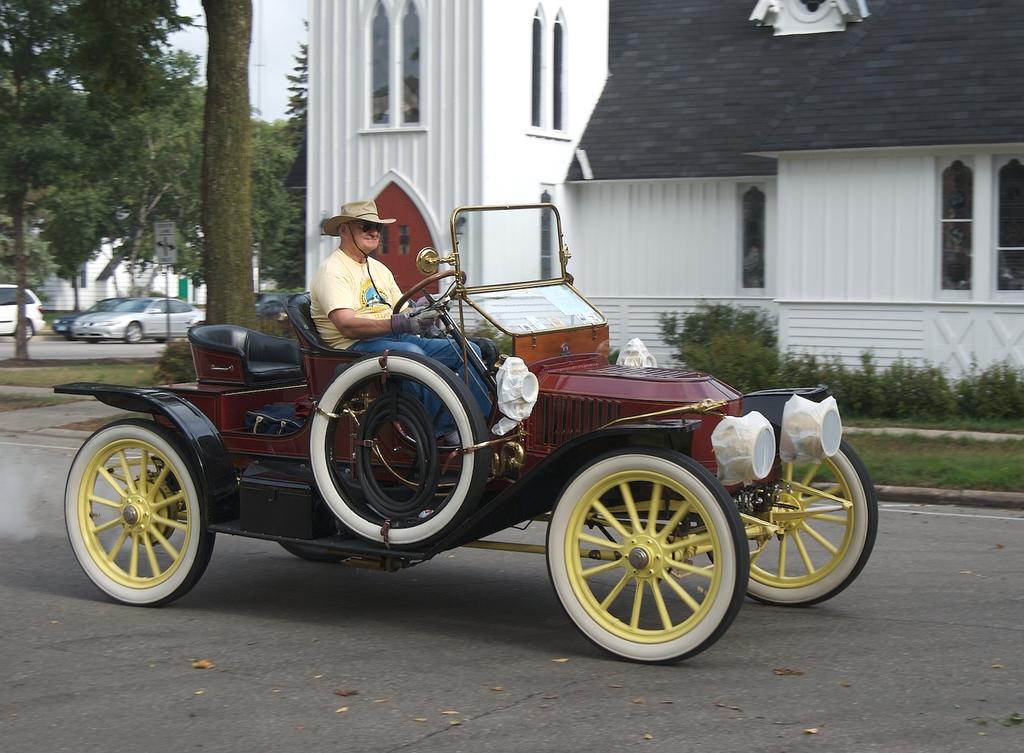}}&
\subfigure{\includegraphics[width=0.065\textwidth]{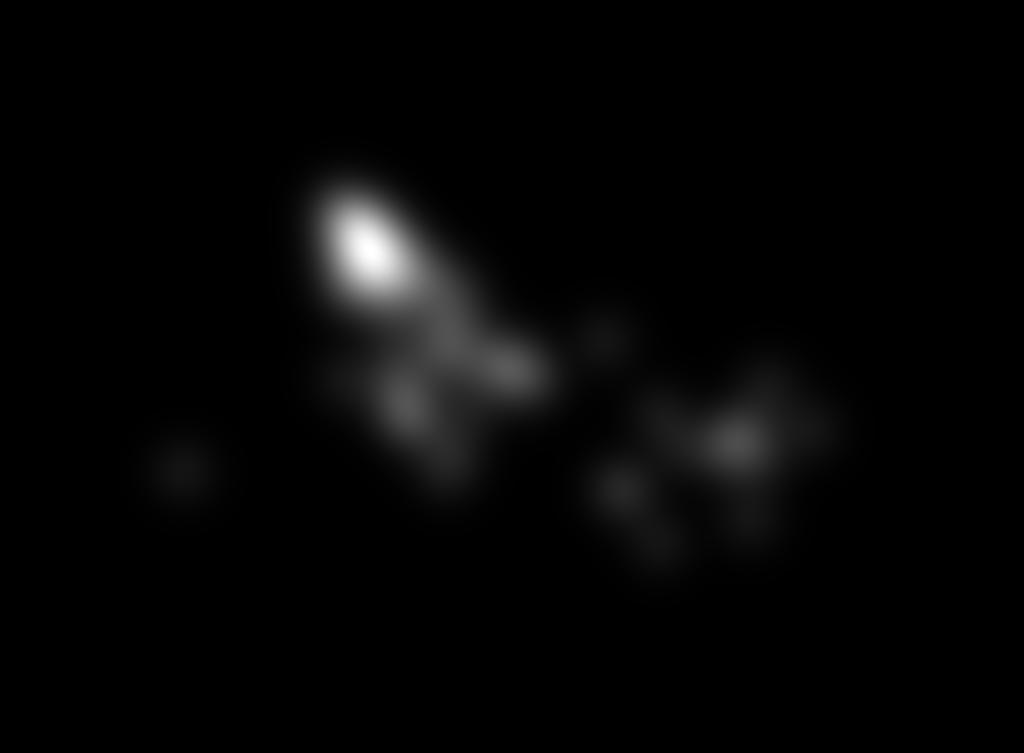}}&
\subfigure{\includegraphics[width=0.065\textwidth]{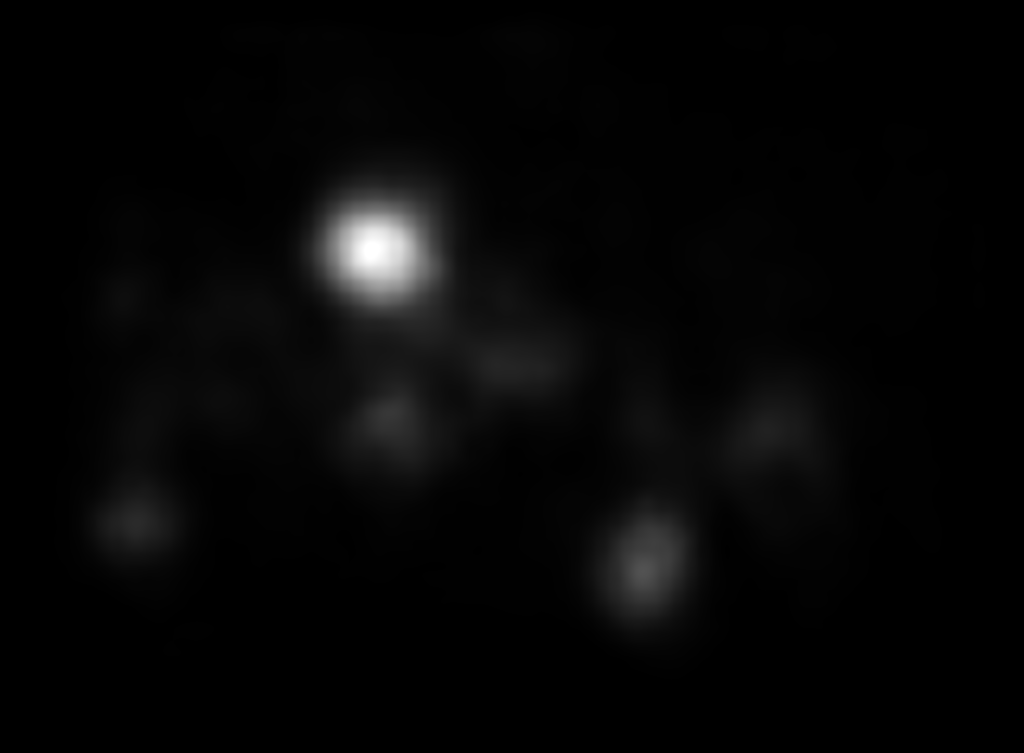}}&
\subfigure{\includegraphics[width=0.065\textwidth]{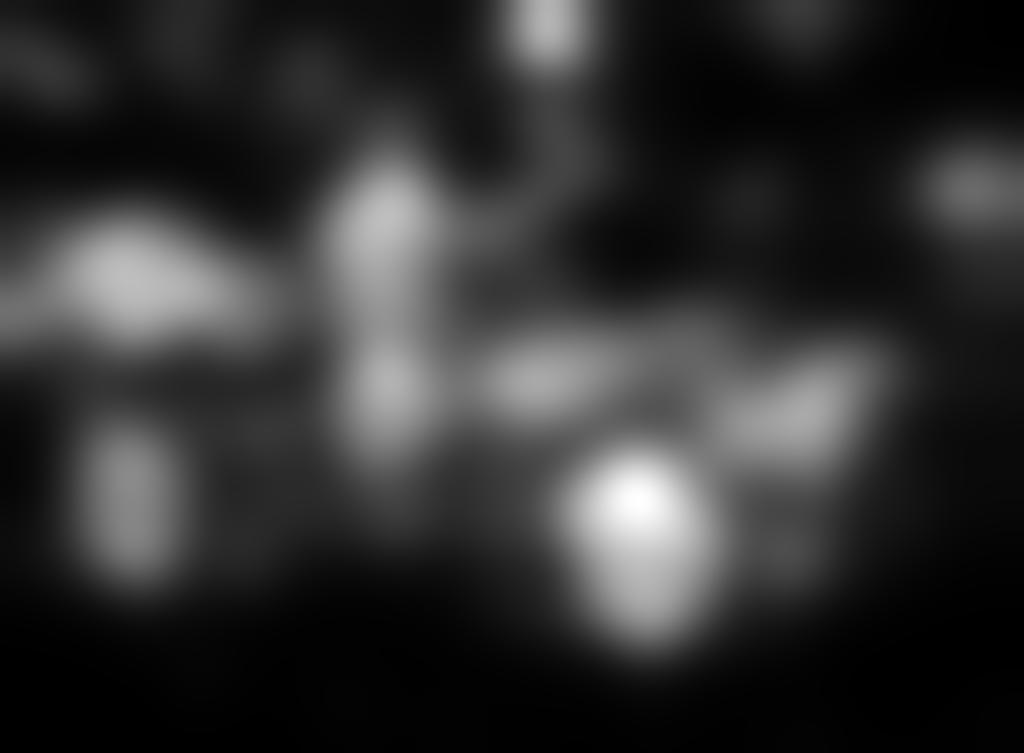}}&
\subfigure{\includegraphics[width=0.065\textwidth]{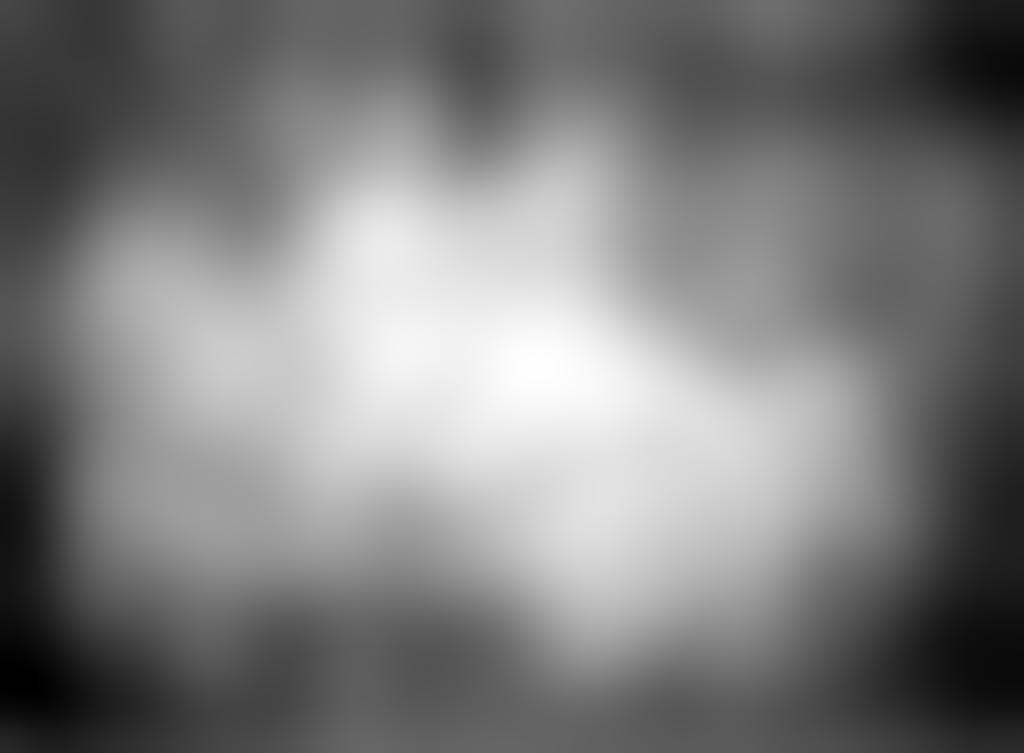}}&
\subfigure{\includegraphics[width=0.065\textwidth]{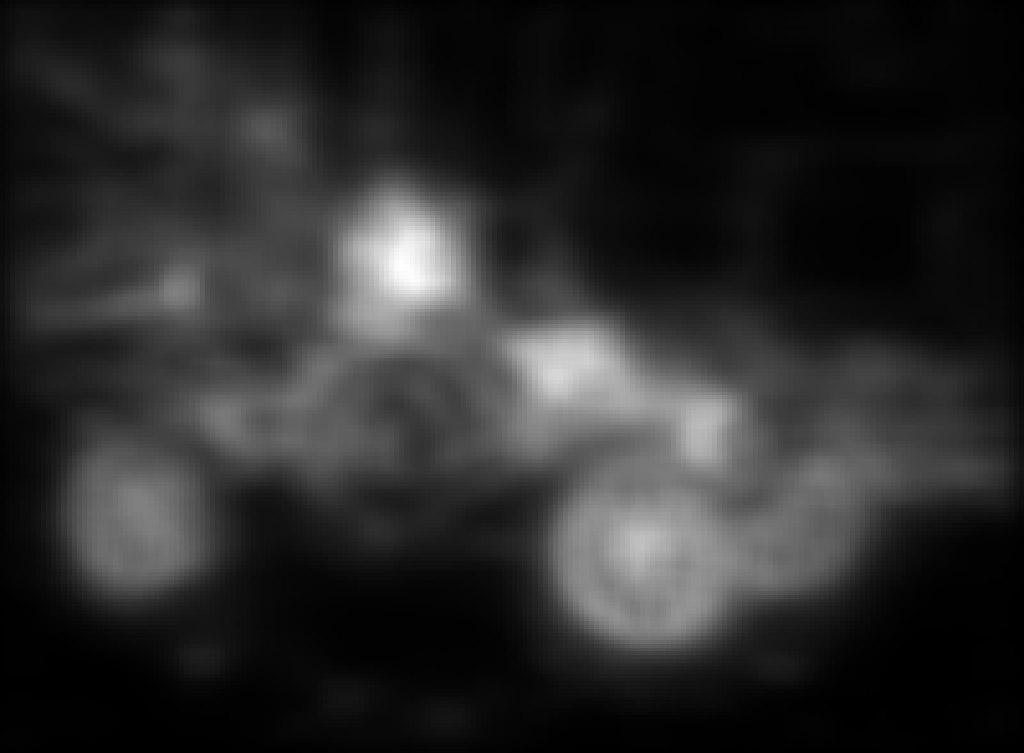}}&
\subfigure{\includegraphics[width=0.065\textwidth]{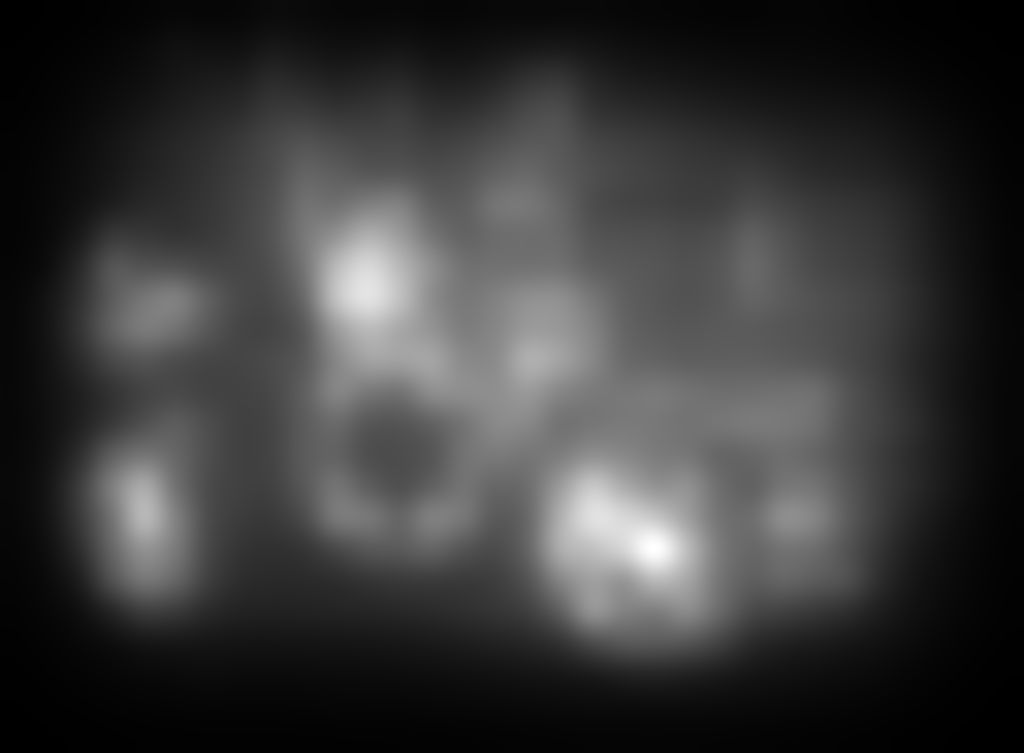}}\\

\end{tabular}
\caption{Qualitative results and comparison to the state of the art. Left: validation images from SALICON dataset~\cite{jiang2015salicon}. Right: validation images from MIT1003 dataset~\cite{judd2009learning}. Best viewed in color.}
\label{fig:res_salicon}
\end{figure*}

Figure~\ref{fig:res_salicon} presents instead a qualitative comparison showing eight randomly chosen input images from SALICON and MIT1003 datasets, their corresponding ground truth annotations and predicted saliency maps. These examples clearly show how our approach is able to predict saliency maps that are very similar to the ground truth, while saliency maps generated by other methods are far less consistent with the ground truth.

\begin{figure}[htbp]
\centering
\setlength\tabcolsep{1.2pt}
\begin{tabular}{cccccc}
\footnotesize{Image} & \footnotesize{GT} & \footnotesize{Our} & \footnotesize{Image} & \footnotesize{GT} & \footnotesize{Our}  \\
\subfigure{\includegraphics[width=0.065\textwidth]{COCOval2014000000027789.jpg}}&
\subfigure{\includegraphics[width=0.065\textwidth]{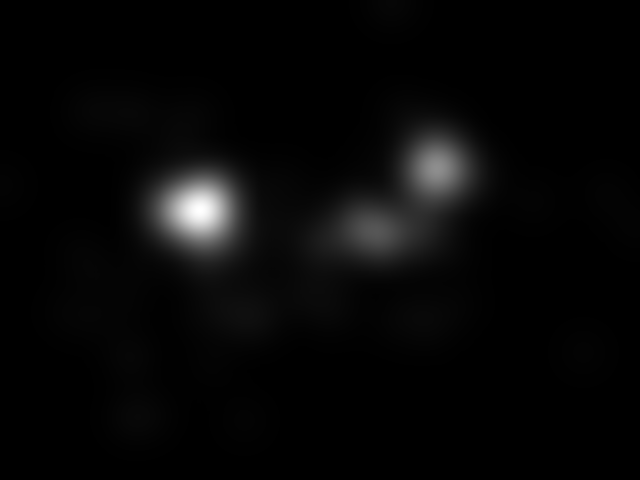}}&
\subfigure{\includegraphics[width=0.065\textwidth]{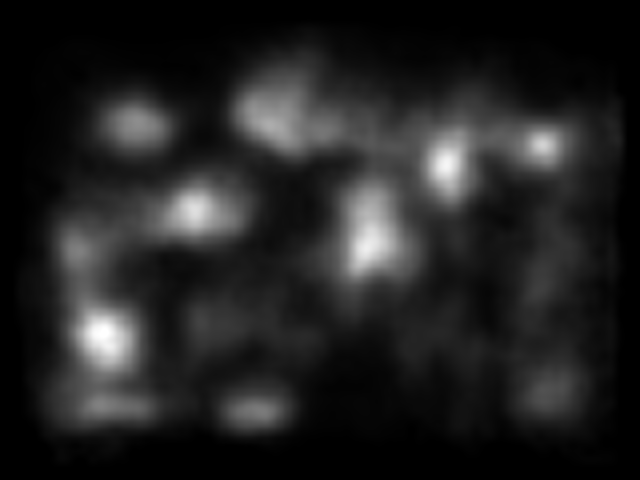}}&
\subfigure{\includegraphics[width=0.065\textwidth]{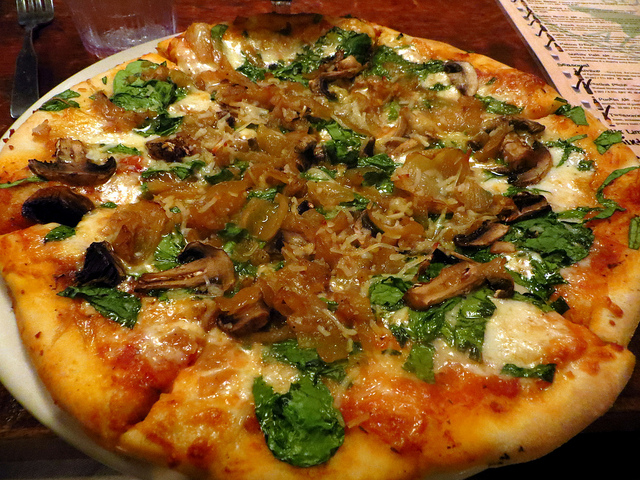}}&
\subfigure{\includegraphics[width=0.065\textwidth]{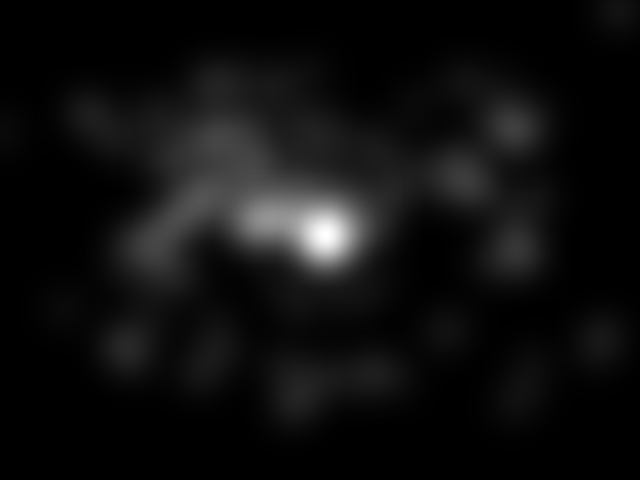}}&
\subfigure{\includegraphics[width=0.065\textwidth]{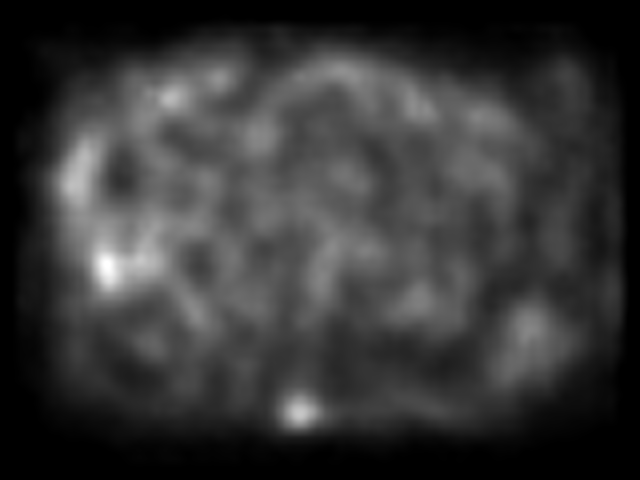}}\\
\subfigure{\includegraphics[width=0.065\textwidth]{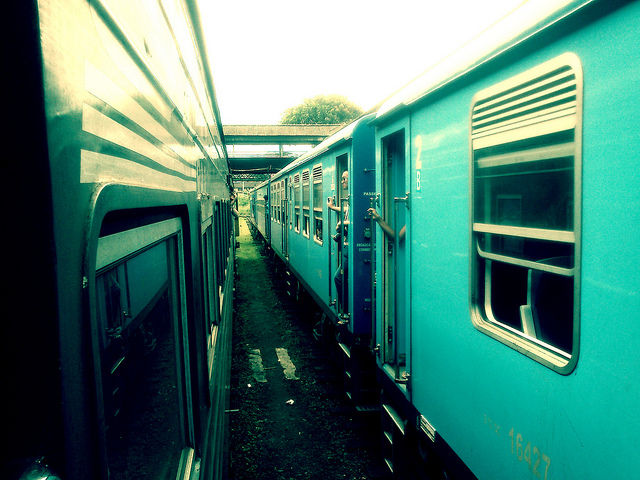}}&
\subfigure{\includegraphics[width=0.065\textwidth]{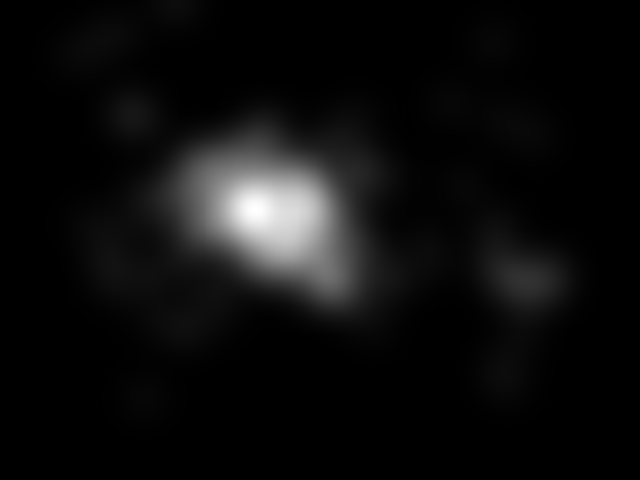}}&
\subfigure{\includegraphics[width=0.065\textwidth]{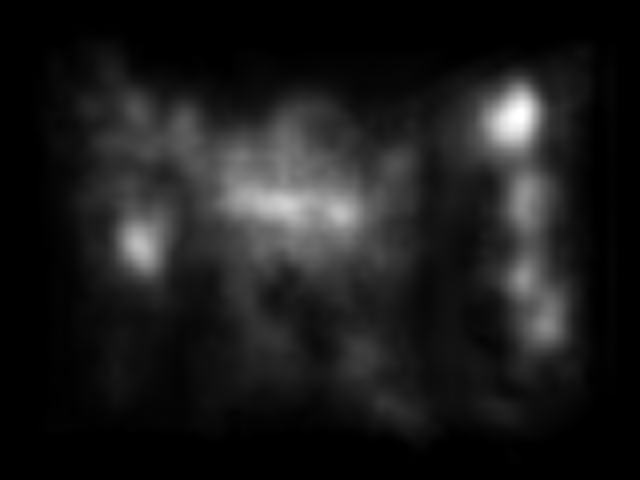}}&
\subfigure{\includegraphics[width=0.065\textwidth]{COCOval2014000000361527.jpg}}&
\subfigure{\includegraphics[width=0.065\textwidth]{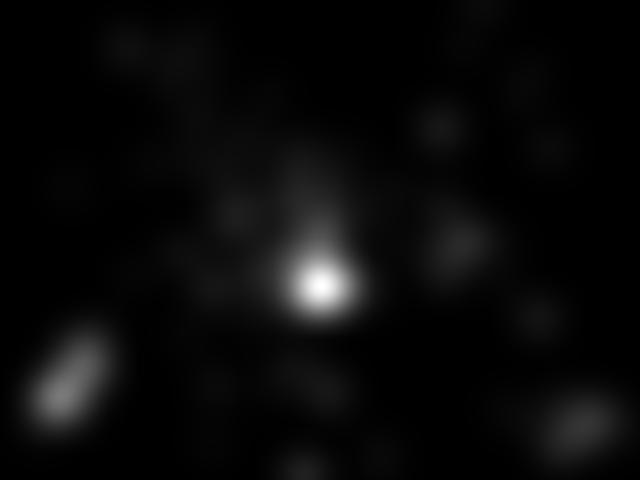}}&
\subfigure{\includegraphics[width=0.065\textwidth]{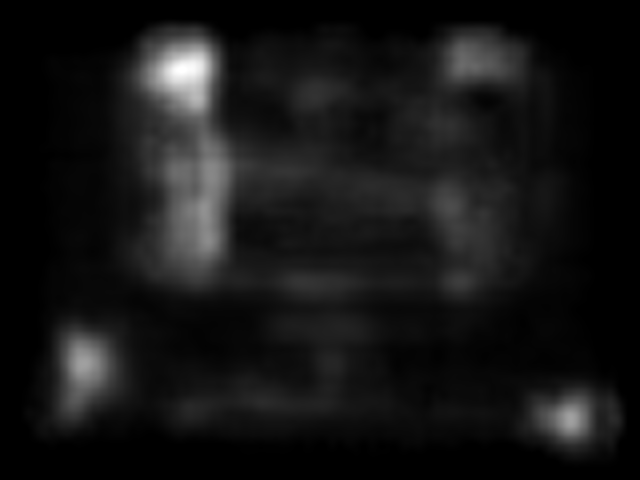}}\\
\end{tabular}
\caption{Example of failure cases on validation images from SALICON dataset~\cite{jiang2015salicon}. Best viewed in color.}
\label{fig:failure}
\end{figure}

Finally, we present some failure cases in Figure~\ref{fig:failure}. As shown, when there is no a clear and explicit object in the image, eye fixations tend to be biased toward image center, which our model fails to predict.

\section{Conclusions}
This paper presented a novel deep learning architecture for saliency prediction. Our model learns a non-linear combination of medium and high level features extracted from a CNN, and a prior to apply to predicted saliency maps, while still being trainable end-to-end. Qualitative and quantitative comparisons with state of the art approaches demonstrate the effectiveness of the proposal on the biggest dataset and on the most popular public benchmark for saliency prediction. 


\section*{Acknowledgment}
We acknowledge the support of NVIDIA Corporation with the donation of the GPUs used in this work.
This work was partially supported by the Fondazione Cassa di Risparmio di Modena project: ``Vision for Augmented Experience'' and the PON R\&C project DICET-INMOTO (Cod. PON04a2 D).

\bibliographystyle{IEEEbib}
\bibliography{bibliography}
\end{document}